\documentclass[journal]{IEEEtran}
\usepackage{graphicx}  

\usepackage{comment}
\usepackage{amsmath,amssymb} 
\usepackage{color}
\usepackage{epsfig}
\usepackage{subfigure}
\usepackage{booktabs}
\usepackage{multirow}
\usepackage{array}
\usepackage{setspace}
\usepackage{mathrsfs}
\usepackage{algorithm}
\usepackage{algorithmic}
\usepackage{caption}

\usepackage{setspace}
\newcommand{\tabincell}[2]{\begin{tabular}{@{}#1@{}}#2\end{tabular}}

\ifCLASSINFOpdf
\else
\fi
\begin{document}
%
\title{Geometry-based Occlusion-Aware Unsupervised Stereo Matching for Autonomous Driving}
%
%
%
%

\author{{Liang Peng,}
        {Dan Deng,}
         {and Deng Cai}}

\IEEEtitleabstractindextext{%

\begin{abstract}
Recently, there are emerging many stereo matching methods for autonomous driving based on unsupervised learning. Most of them take advantage of reconstruction losses to remove dependency on disparity groundtruth. Occlusion handling is a challenging problem in stereo matching, especially for unsupervised methods. Previous unsupervised methods failed to take full advantage of geometry properties in occlusion handling. In this paper, we introduce an effective way to detect occlusion regions and propose a novel unsupervised training strategy to deal with occlusion that only uses the predicted left disparity map, by making use of its geometry features in an iterative way. In the training process, we regard the predicted left disparity map as pseudo groundtruth and infer occluded regions using geometry features. The resulting occlusion mask is then used in either training, post-processing, or both of them as guidance. Experiments show that our method could deal with the occlusion problem effectively and significantly outperforms the other unsupervised methods for stereo matching. Moreover, our occlusion-aware strategies can be extended to the other stereo methods conveniently and improve their performances.
\end{abstract}

}

\maketitle

\IEEEdisplaynontitleabstractindextext

%
\IEEEpeerreviewmaketitle

\section{Introduction}\label{sec:introduction}

	\IEEEPARstart{S}{tereo} matching is a classical problem in computer vision. It has been widely used in many fields, especially for autonomous driving. Given a pair of rectified left and right images, the displacement of corresponding matching pixels is called disparity. Depth can be derived by disparity according to triangulation.
	
	Benefiting from the development of Convolutional Neural Networks(CNNs), many deep learning methods show amazing results in stereo matching. However, most of these methods require massive high-quality groundtruth of disparities for training, which is expensive to obtain. And therefore, many unsupervised-learning-based methods have been proposed, mainly based on image reconstruction losses \cite{Godard,Zhou,UnOS}. Such losses evaluate the differences between the reconstructed left image and the original left image via similarity metrics, e.g., Structure Similarity (SSIM)\cite{SSIM}, L1.  The reconstruction process usually relies on the differentiable warping mechanism similar to Spatial Transformer Network (STN)\cite{STN}. Pixels on the left image are recovered by bilinear sampling from the original right image according to the corresponding disparity values.

	\begin{figure}[t]
		\begin{center}
			\begin{minipage}[t]{1\linewidth}
				\centering
				\includegraphics[width=1\linewidth]{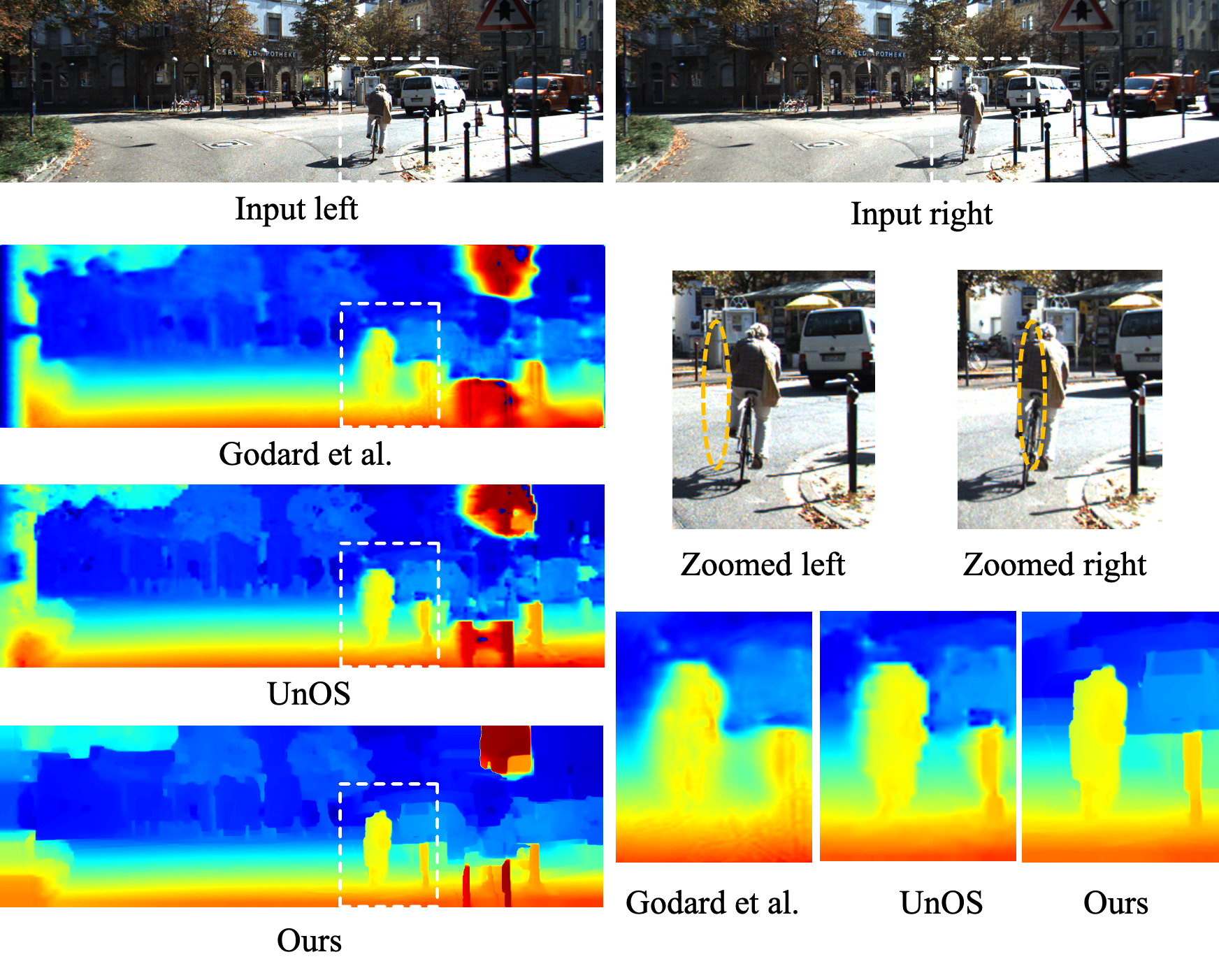}
			\end{minipage}
			
			
		\end{center}
		\caption{Comparison examples of disparity prediction results from KITTI 2015\cite{KITTI_2015}. Regions in the dashed ellipses of the zoomed left image are occluded in the zoomed right image by the bicycler. Among the compared state-of-the-art unsupervised stereo methods\cite{Godard,UnOS}, ours performs the best, especially on occluded regions. Best viewed in color.}
		\label{fig:comparision}
		\label{fig:onecol}
	\end{figure}

However, in binocular settings, regions visible in one camera may be unseen in another. It is difficult or even impossible to reconstruct such pixels in one image from the other. And therefore, the reconstruction losses calculated on these pixels are meaningless and noisy, imposing a negative impact on model performances, especially on occluded regions. As observed in experiments, disparity images inferred by unsupervised methods tend to be blurred and indistinguishable from their surrounding areas in occlusion regions.  Some examples are shown in Fig. \ref{fig:comparision}. Recent unsupervised methods usually adopt the left-right consistency check\cite{Godard,Zhou,SegStereo,UnOS} or a CNN\cite{Li} to detect or predict occlusions. These methods ignore or fail to make full use of the geometry properties of occlusions.

	\begin{figure*}[htbp]
		\begin{center}
			
			\subfigure[]{
				\includegraphics[width=0.45\linewidth]{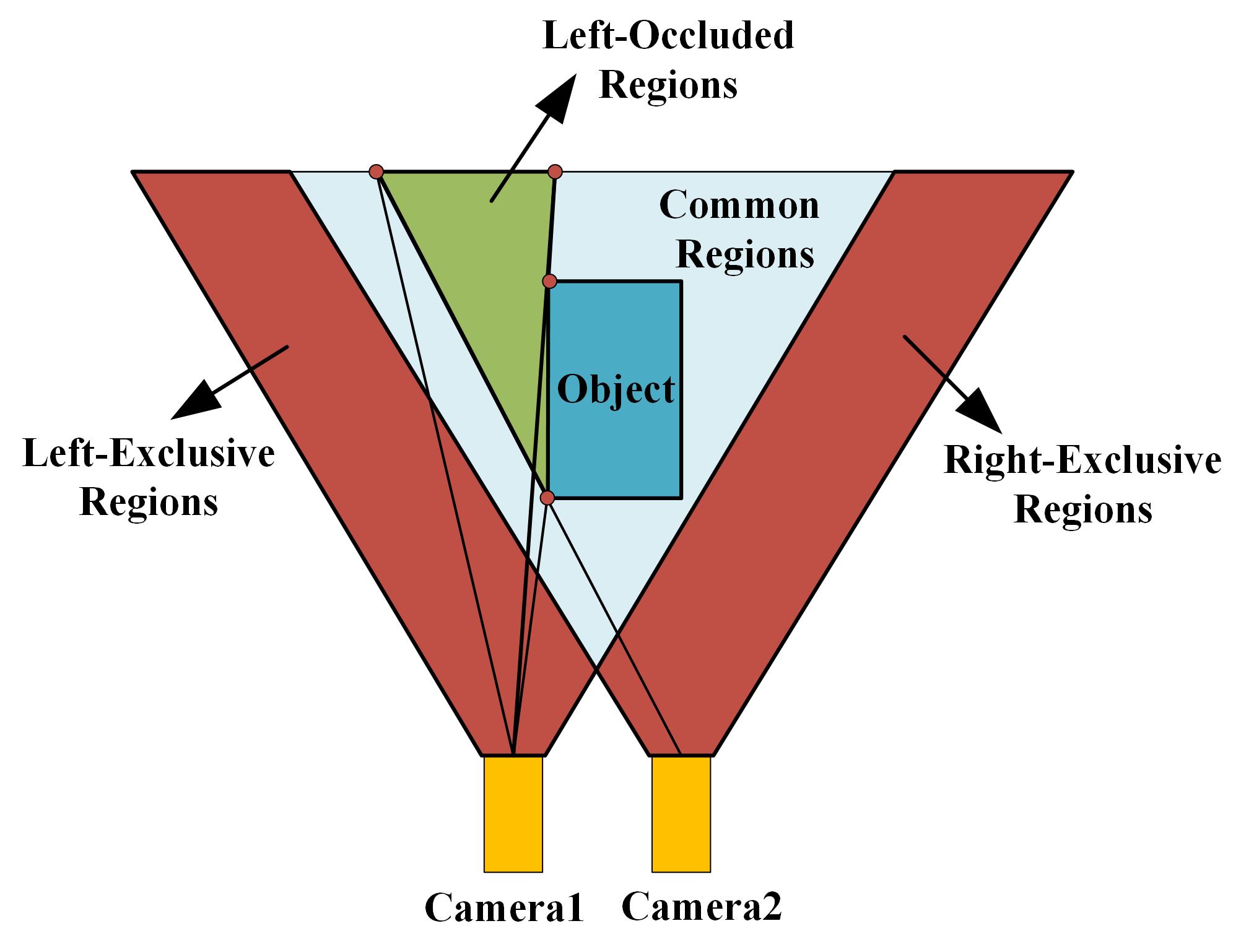}
			}
			\subfigure[]{
				\includegraphics[width=0.45\linewidth]{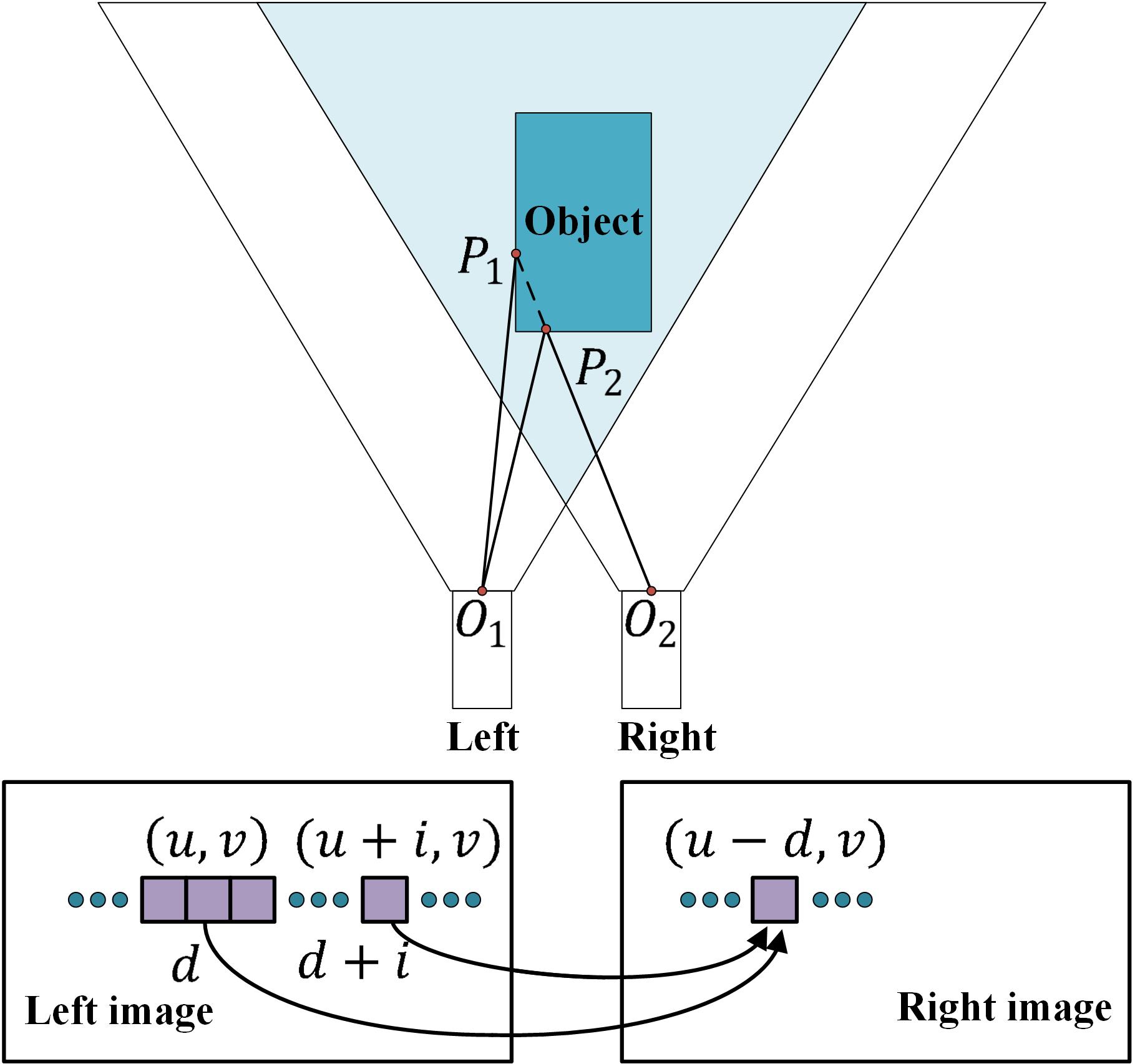}
			}
			
		\end{center}	
		\caption{Occlusion illustration. The Left and Right are the two cameras in a typical binocular system. The Object is opaque to light. (a): Definitions of different types of regions in a typical binocular system. (b): Illustration of the condition for a point to be occluded. $O_1, O_2$ are the optical centers of the cameras. $d$ denotes disparity. $P_1$ and $P_2$ in the left image are mapped to a same point of right image, so $P_1$ is occluded by $P_2$.}
		\label{fig:occlusion-illustration}
		
		\label{fig:onecol}
	\end{figure*}

In this paper, we propose Geometry-based Occlusion-Aware Unsupervised Stereo Matching(GOUSM), a novel occlusion-aware method. The disparity prediction from the  model being trained is used as pseudo groundtruth, from which occluded regions can be located based on epipolar geometry constraints and represented as an occlusion mask. It is then applied in the training process to mask the reconstruction loss calculated in occluded regions. The whole process is carried out iteratively epoch by epoch, and the quality of disparity predictions and occlusion masks are improved gradually.  The occlusion mask is also used in our post-processing to refine predictions in occluded regions. Experiments show that, by taking advantage of occlusion geometry constraints, our method could deal with occlusion more effectively, and significantly outperforms the other unsupervised stereo matching methods. Moreover, our approaches can be easily extended to other stereo methods and improve their performances.
	
	Our main contributions can be summarized as below:
	\begin{enumerate}
		\item We  propose a novel and effective geometry-based approach to deal with the occlusion problem for unsupervised stereo matching.
		\item Our method can be adopted conveniently by the other stereo methods and helps to improve their performances.
		\item Our method achieves state-of-the-art performances, significantly outperforming the other unsupervised learning methods.
	\end{enumerate}

	\section{Related Work}
	
	\subsection{Supervised Stereo Matching}
	Early supervised methods focus on accurately computing the matching cost with CNNs and refining the disparity map with semi-global matching (SGM)\cite{SGM}, such as \cite{C1,C2,C3}. Inspired by pixel-wise labeling tasks, Mayer et al.\cite{DispNet} developed end-to-end learning of disparity maps by using Fully-Convolution Network(FCN\cite{FCN}). Chang proposed PSMNet\cite{PSMNet}, getting much better results by taking advantage of pyramid spatial pooling and hourglass network, which are time-consuming. To achieve real-time inference, Khamis et al. proposed StereoNet\cite{StereoNet}, with small network architecture and high speed. Most supervised methods \cite{DispNet,PSMNet,Corr2,GC-Net,Corr3,Corr1,CRL} build cost volumes by concatenation or multiplication. GwcNet\cite{GWC} is proposed to calculate cost volume by group-wise correlation to balance speed and accuracy. High-quality groundtruth is indispensable for supervised methods. However, it's expensive to yield enough.

	\subsection{Unsupervised Stereo Matching}
	Unsupervised stereo matching is becoming popular recently because it does not need groundtruth. Chao et al.\cite{Zhou} proposed an iterative unsupervised learning network that uses left and right checks to select the appropriate point matching pair, which is then cycled a certain number of times to optimize. Godard et al.\cite{Godard} proposed the reconstruction mechanism for unsupervised stereo matching, and has been followed by most later unsupervised stereo methods. UnOS\cite{UnOS} employs both consequent frames and stereo pairs to jointly learn optical flow and disparity maps. Some work takes the unsupervised loss to get better generalization, such as MADNet\cite{MADNet}. It uses the unsupervised loss for online adaptation.
	
	\subsection{Occlusion-Aware Methods}
	To tackle occlusions in stereo matching, Fua\cite{Fua} created two disparity maps relative to each image: one from left to right and another from right to left to check their consistency, i.e.,left-right consistency check. Many later methods\cite{Sun,Godard,UnOS} follow this strategy. By exploiting the symmetric property of optical flow\cite{Teng,Wang}, this check can also be extended to optical flow for occlusion handling. Zitnick\cite{C} examines the magnitude of the converged match values in conjunction with the uniqueness constraint, and Luo et al.\cite{Luo} rely on continuity. Li et al.\cite{Li} detect occlusions depend on direct model inference, i.e., predicting an occlusion mask by adding special submodules in the CNN model. 


	\begin{figure*}[htbp]
		\centering
		\subfigure[\scriptsize{Input left}]{
			\includegraphics[width=0.23\linewidth]{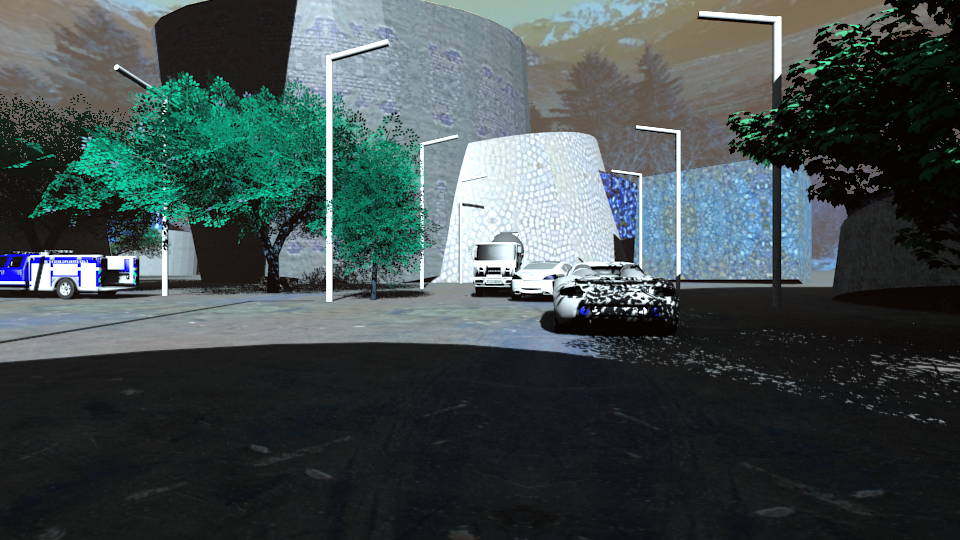}
		}
		\subfigure[\scriptsize{Input right}]{
			\includegraphics[width=0.23\linewidth]{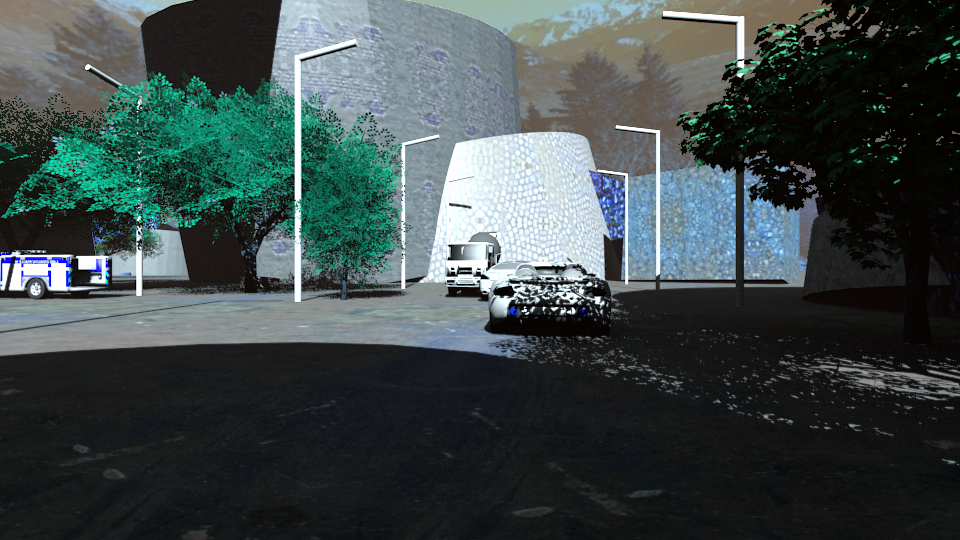}
		}
		\subfigure[\scriptsize{groundtruth left disparity}]{
			\includegraphics[width=0.23\linewidth]{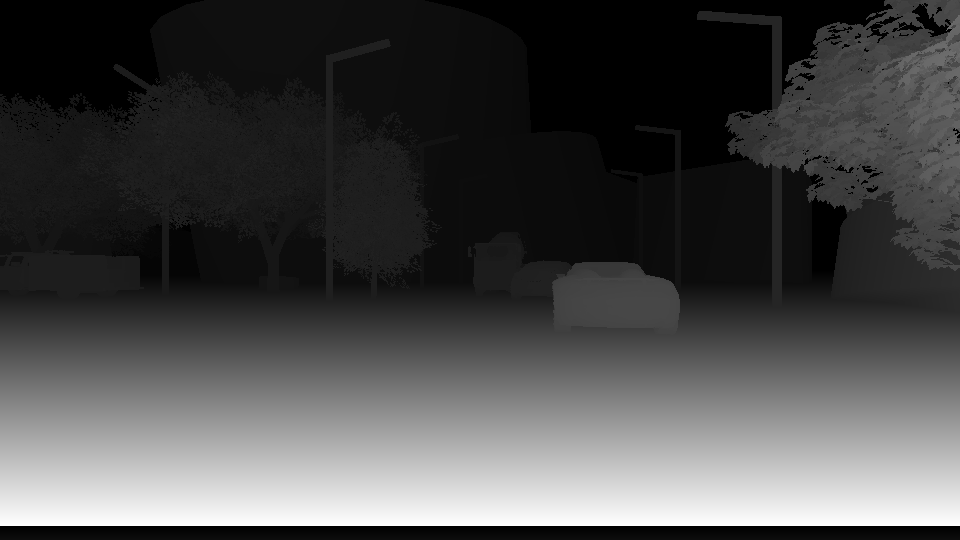}
		}
		\subfigure[\scriptsize{Occlusion mask}]{
			\includegraphics[width=0.23\linewidth]{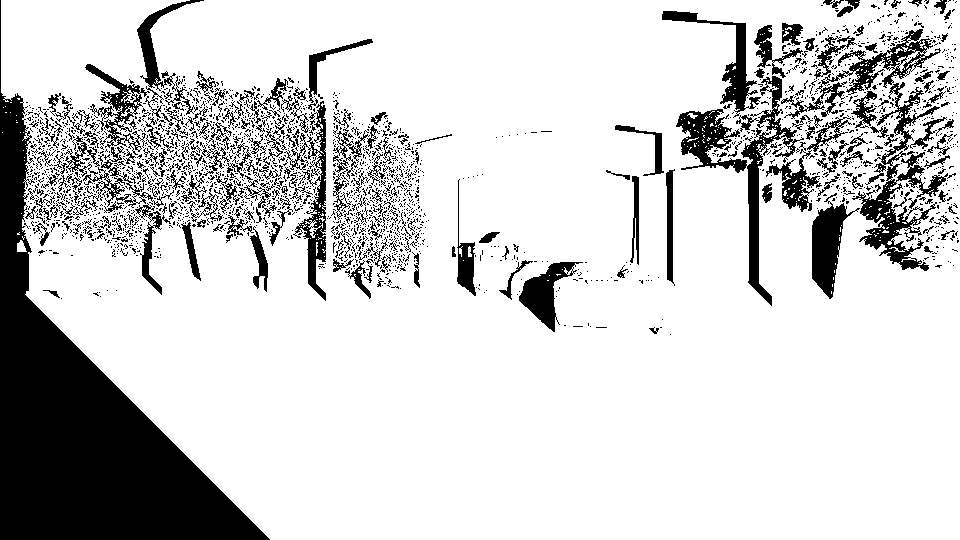}
		}
		\subfigure[\scriptsize{Reconstructed left image, without dealing with occlusion}]{
			\includegraphics[width=0.23\linewidth]{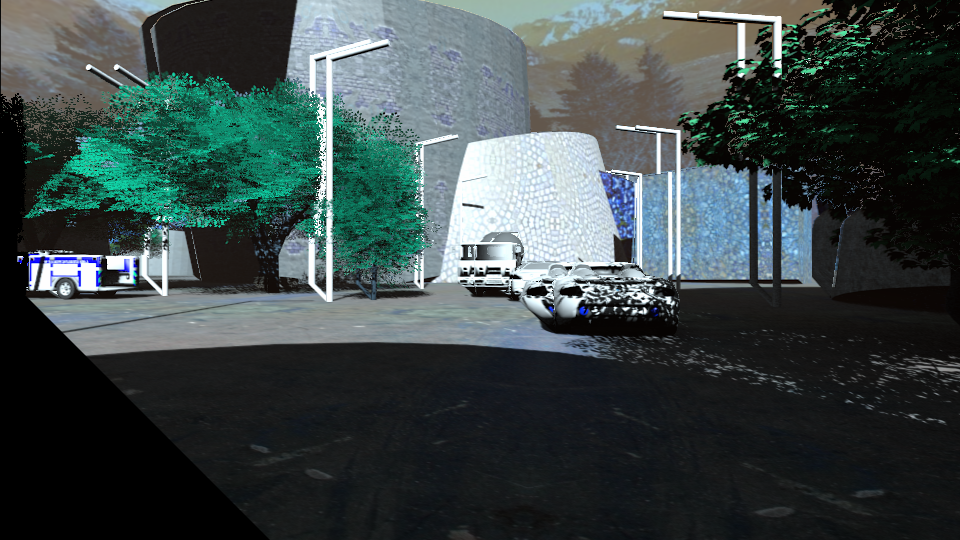}
		}
		\subfigure[\scriptsize{Reconstructed left image after dealing with occlusion}]{
			\includegraphics[width=0.23\linewidth]{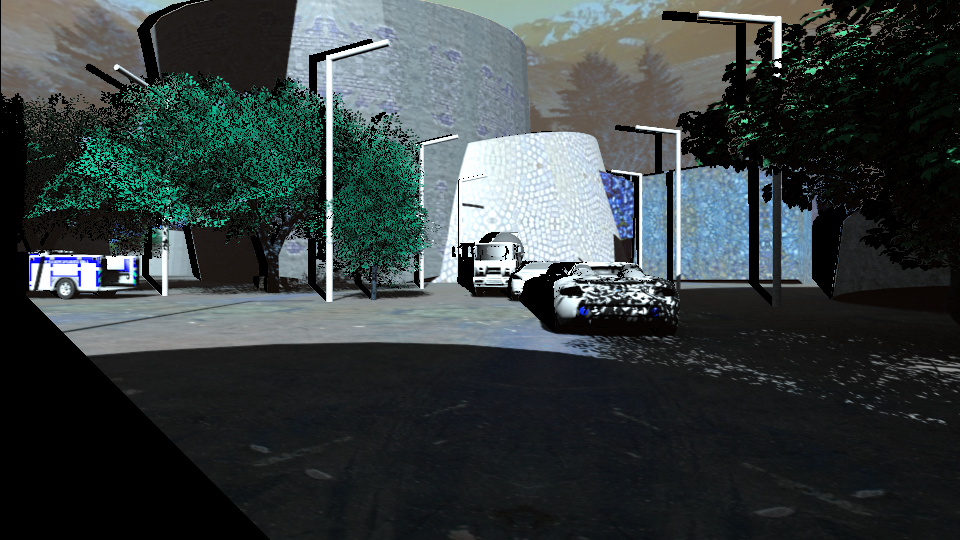}
		}
		\subfigure[\scriptsize{Reconstruction error of $(e)$}]{
			\includegraphics[width=0.23\linewidth]{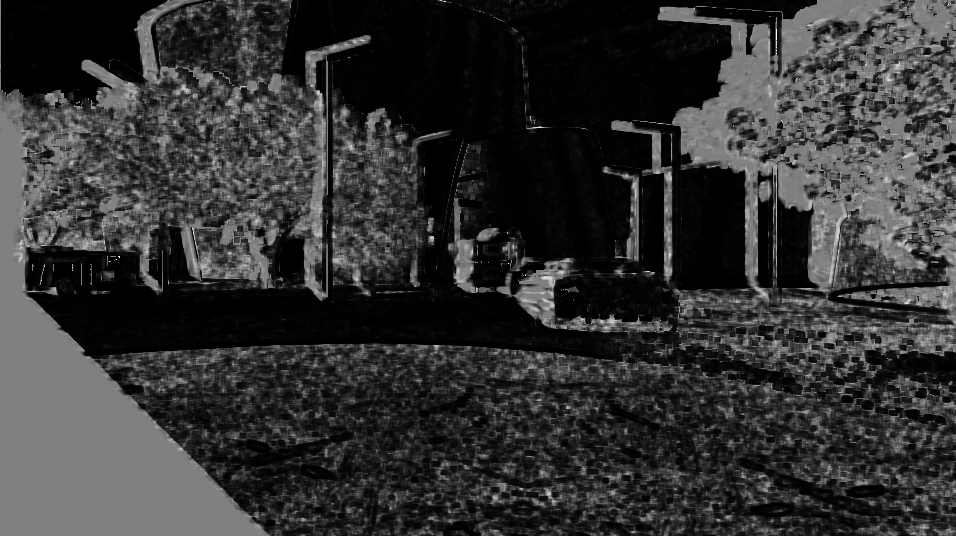}
		}
		\subfigure[\scriptsize{Reconstruction error of $(f)$}]{
			\includegraphics[width=0.23\linewidth]{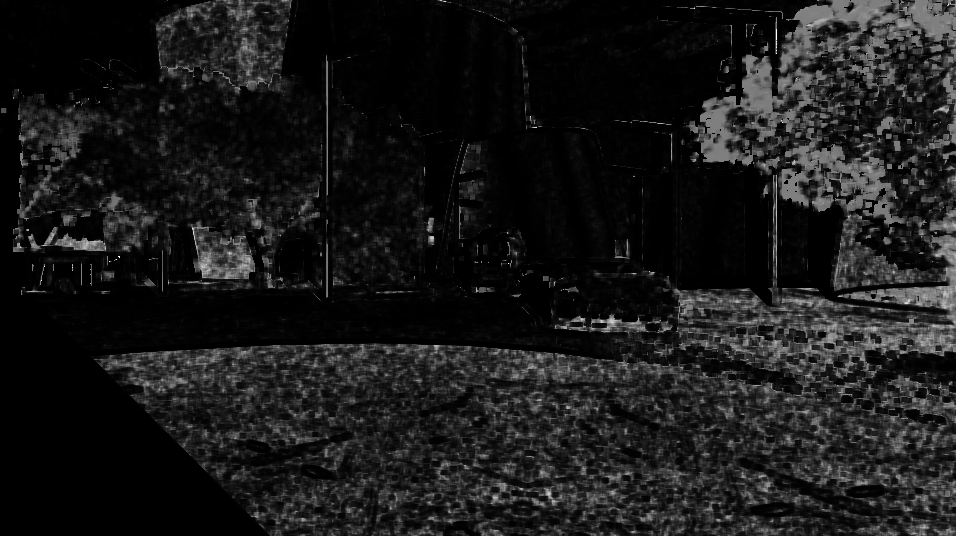}
		}
		\caption{Illustration of the influences of occlusion in the reconstruction process. The occlusion mask is derived only from left disparity. The black regions represents occlusion in (d). In the error images (g) and (h), brighter color mean larger error. It can be seen that reconstruction without being occlusion-aware produces a high error in occlusion regions.}
		\label{fig:sceneflow}
		\end{figure*}

	\section{Occlusion Analysis}
		
	\subsection{Occlusion Definition and Detection}
	Before elaborating on the whole pipeline, we will show how to detect occluded regions by geometry and their influences on model training. Some definitions are declared at first for a better explanation. As illustrated in Fig. \ref{fig:occlusion-illustration}(a),
	\begin{enumerate}
		\item Common Regions, the intersection of left and right viewing frustums.
		\item Exclusive Regions, the regions not in Common Regions.
		\item Occluded Regions, the regions in Common Regions only visible in one camera.
	\end{enumerate}
	
	All occlusions we mentioned in the following sections refer to occluded regions of the left camera, i.e., visible in left but unseen from right, if not specified.

	Then, we will derive the condition for a given point to be occluded. 
	As is illustrated in Fig. \ref{fig:occlusion-illustration}(b), $O_1 = (0, 0, 0), O_2 = (b, 0, 0)$ are the optical centers of left and right cameras  in a binocular system after rectification\cite{Rectified}, respectively. The "Object" is opaque to light. $P_1$ is a 3D point on the object surface. $P_2$ is the crosspoint where straight line $P_1O_2$ intersects with the object surface.  In the imaging process, $P_1$ and $ P_2$ are both visible in the left image. However, since $P_1$ is occluded by $P_2$, it is invisible from the right. 
	
	Denote the projection points of $P_1, P_2$ on the left image as  
	\begin{equation}
	p_1 = (u_1, v_1),
	\end{equation}
	\begin{equation}
	p_2= (u_2, v_2),  
	\end{equation}
	and their disparities as $d_1, d_2$, respectively. Their projection points on the right image can be written as 
	\begin{equation}
	p_1' = (u_1 - d_1, v_1),
	\end{equation}
	\begin{equation}
	p_2' = (u_2 - d_2, v_2).
	\end{equation}
	Because $P_1, P_2, O_2$ are all in the same straight line, $P_1$ and $P_2$ are mapped to the same pixel on the right image,  then
	\begin{equation}
	u_1 - d_1 = u_2 - d_2,
	\end{equation}
	\begin{equation}
	v_1 = v_2.
	\end{equation}
	Let $\Delta u = u_2 - u_1$, $\Delta d = d_2 - d_1$, we get $\Delta u = \Delta d$ for the occluded point $P_1$.
	
	It can be further concluded that \textbf{for a given pixel $\mathbf{p_1}$ in the left image, it belongs to Left-Occluded Regions when there exists another pixel $\mathbf{p_2}$ satisfying}
	\begin{equation}
	\mathbf{\Delta u = \Delta d > 0}.
	\label {equ:occlusion-condition}
	\end{equation}

	Furthermore, if
	\begin{equation}
	\mathbf{ d_1 -  u_1 > 0},
	\label {equ:occlusion-condition2}
	\end{equation}
     \textbf{$\mathbf{p_1}$ belongs to Left-Exclusive Regions} since it is already beyond the FOV (Field Of View)  of the right camera.
	
	Given a disparity image, occluded pixels can be located via Eq. \ref{equ:occlusion-condition} and Eq. \ref{equ:occlusion-condition2}, and represented as a mask, namely, occlusion mask.

	\subsection{The Influences of Occlusion on Unsupervised Stereo Methods}
	In binocular systems, occluded regions exist extensively.  Unsupervised stereo matching methods mainly depend on losses calculated from reconstruction in training. However, given the disparity and right images, it is difficult or even impossible to reconstruct the occluded regions since they are invisible on the right image. And therefore, the reconstruction losses calculated on these pixels are meaningless and noisy, imposing a negative impact on model performances, especially on occluded regions. 
	
	As is shown in Tab. \ref{tab-occlusion-stat} and Fig. \ref{fig:sceneflow},  occluded regions occupy about 20\% on left images from SceneFlow\cite{DispNet},  in average.  The total L1 reconstruction error from occlusion regions is 2.32 times as that on non-occlusion.  Consequently, the impact of occlusion is too significant to be ignored for unsupervised stereo matching methods.  As shown in Fig. \ref{fig:comparision}, disparity images inferred by the state-of-the-art unsupervised methods tend to be blurred and indistinguishable from their surrounding pixels in occluded regions.

		\begin{table}
		\label{tab-occlusion-stat}
		\caption{Occlusion statistical on SceneFlow\cite{DispNet}. SceneFlow is a synthesized dataset containing 35,454 pairs of stereo images with precise and dense disparity groundtruth. L1 error is calculated on the whole dataset. Mean Error: mean L1 error per pixel; Total error: the proportion of error from all the pixels from different regions w.r.t. the total error from all regions; Area: the proportion of the areas of different regions w.r.t. total area of all regions.}
		\begin{center}
			\setlength{\tabcolsep}{2mm}{
				\begin{tabular}{cccc}
					\toprule  
					& Mean Error & Total Error & Area\\ 
					\midrule         
					Occluded & 39.78 & 69.88\% & 16.86\% \\
					Non-Occluded & 3.47 & 30.12\% & 83.14\% \\
					\multirow{2}{*}{$\frac{Occluded}{Non-occluded}$} &\multirow{2}{*}{11.46} & \multirow{2}{*}{2.32} & \multirow{2}{*}{0.20}\\
					~&~&~&~\\
					\bottomrule 
			\end{tabular}}
		\end{center}
		\label{tab-occlusion-stat}
	\end{table}
	
	\begin{figure}[t]
		\begin{center}
			\begin{minipage}[t]{1\linewidth}
				\centering
				\includegraphics[width=1\linewidth]{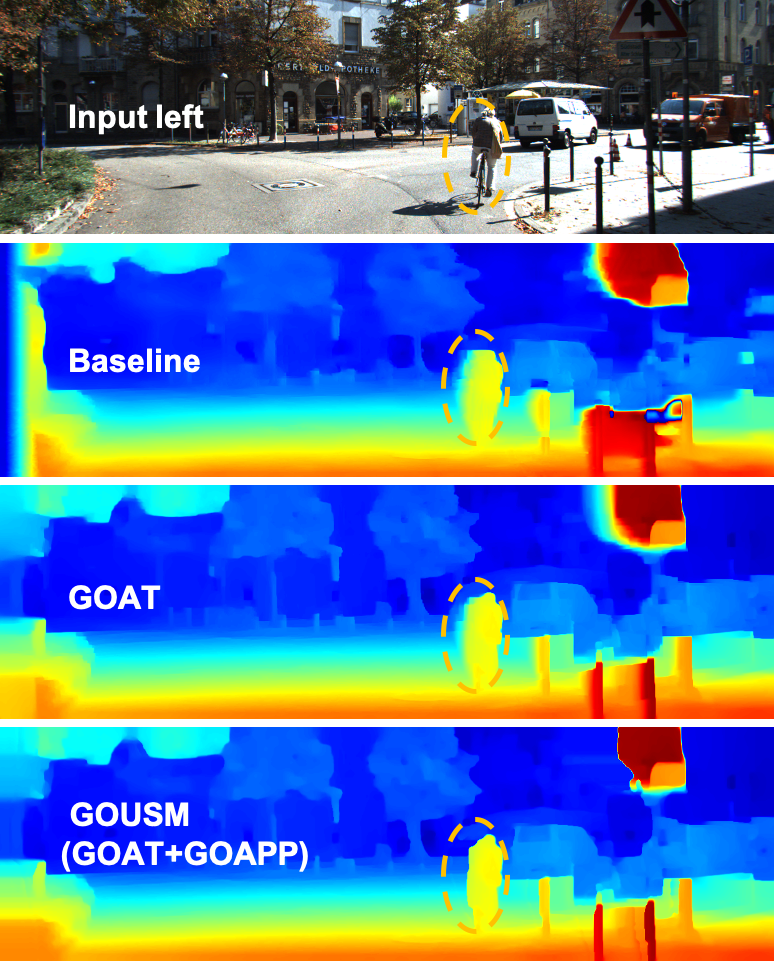}
			\end{minipage}
		\end{center}
		\caption{Qualitative results of GOAT and GOAPP. By employing GOAT and GOAPP, it can be observed that the result is gradually less affected by occlusions and retain more details. Best viewed in color.}
		\label{fig:post-processing}
		
		\label{fig:onecol}
	\end{figure}

	\section{Geometry-based Occlusion-Aware Unsupervised Stereo Matching}
	
	\subsection{Occlusion-Aware Pipeline}
	\subsubsection{Gometry-based Occlusion-Aware Training (GOAT)}
   Occlusions would impose a negative impact on unsupervised stereo matching methods if not properly processed during training. A basic and intuitive solution is to find the occluded pixels via Eq. \ref{equ:occlusion-condition} and Eq. \ref{equ:occlusion-condition2}, then ignore them in loss calculation. However, groundtruths of disparities are supposed to be absent in unsupervised training, while accurate occlusion detection relies on precise disparity and vice versa. In a word, accurate occlusion first or accurate disparity first is a chicken-and-egg problem. Our method deals with this contradiction in an iterative way. We use disparity predictions to find occlusions and ignore them in the training process. As is listed in Alg.\ref{alg:training}, after the first epoch trained on all pixels, occlusion masks are updated offline after each epoch, instead of each iteration, to make the training more stable and efficient in convergence. At first, the predicted disparities and occlusion masks might be noisy, but their precisions would increase as the training goes on. 
   
	\begin{algorithm}[htb] 
		\caption{GOAT:Geometry-based Occlusion-Aware Training} 
		\begin{algorithmic}[1] 
			\STATE Initial model parameters
			\STATE Train on all pixels in the 1st epoch
			\FOR{$i\gets $ from $2$ to $total\hspace{2pt} epochs$}
			\STATE Update occlusion masks of all training samples
			\STATE Train without occluded regions
			\ENDFOR
			\STATE Obtain the final model
		\end{algorithmic}
		\label{alg:training}
	\end{algorithm}
	
	\subsubsection{Geometry-based Occlusion-Aware Post Processing (GOAPP)}
  We believe that disparity predictions for occluded regions tend to be more unreliable than those for non-occlusions, no matter GOAT is adopted or not. And therefore, the occlusion mask is then used in post-processing to refine predictions for occluded pixels. Different type occlusion regions are dealt using different ways via Alg.\ref{alg:post-processing}. The basic idea behind is re-calculating the disparities of occluded pixels from their non-occluded neighbors, by nearest-neighbor sampling and linear interpolation. $N$ is the number of neighbors used in sampling and interpolation and is set to 10 in default.  It's worth noting that GOAPP is much different from the usual post-processing applied in previous methods. On the one hand, GOAPP requires only a single predicted disparity map to detect occluded regions, nevertheless, other methods usually need to obtain both left and right disparity maps or a  CNN, limiting their application scenarios. On the other hand, we adopt different strategies to handle different types of occlusions to be tailored to fit better into the scenes of autonomous driving.
	
	\begin{algorithm}[htb] 
		\caption{GOAPP: Geometry-based Occlusion-Aware Post Processing} 
		\begin{algorithmic}[1] 
			\REQUIRE
			Occlusion mask $M$
			\REQUIRE
			Disparity image $D$
			\FOR{$i \gets$ from 0 to $NumRows(D)$}
			\STATE find $\hspace{2pt} k  \hspace{2pt} $satisfying:$ \hspace{2pt} \hspace{2pt} All(M_{i,j<k})\!=\!0 \hspace{2pt}$ and $\hspace{2pt} M_{i,k+1}\!=\!1$
			\FOR{$j \gets$ from $ k $ to $0$}
			\STATE $D_{i,j}\gets\frac{1}{N}\sum_{m=N+1}^{1}M(i,j+m)$
			\ENDFOR
			
			\FOR{$j \gets$ from $k$ to $NumCols(D)$}
			\STATE $D_{i,j}\gets\frac{1}{N}\sum_{m=-N-1}^{-1}M(i,j+m)$
			\ENDFOR
			\ENDFOR
			\STATE Return $D$
		\end{algorithmic}
		\label{alg:post-processing}
	\end{algorithm}


	\subsection{Network Details.}
	
	\begin{figure*}[t]
		\begin{center}
			\begin{minipage}[t]{0.8\linewidth}
				\centering
				\includegraphics[width=0.7\linewidth]{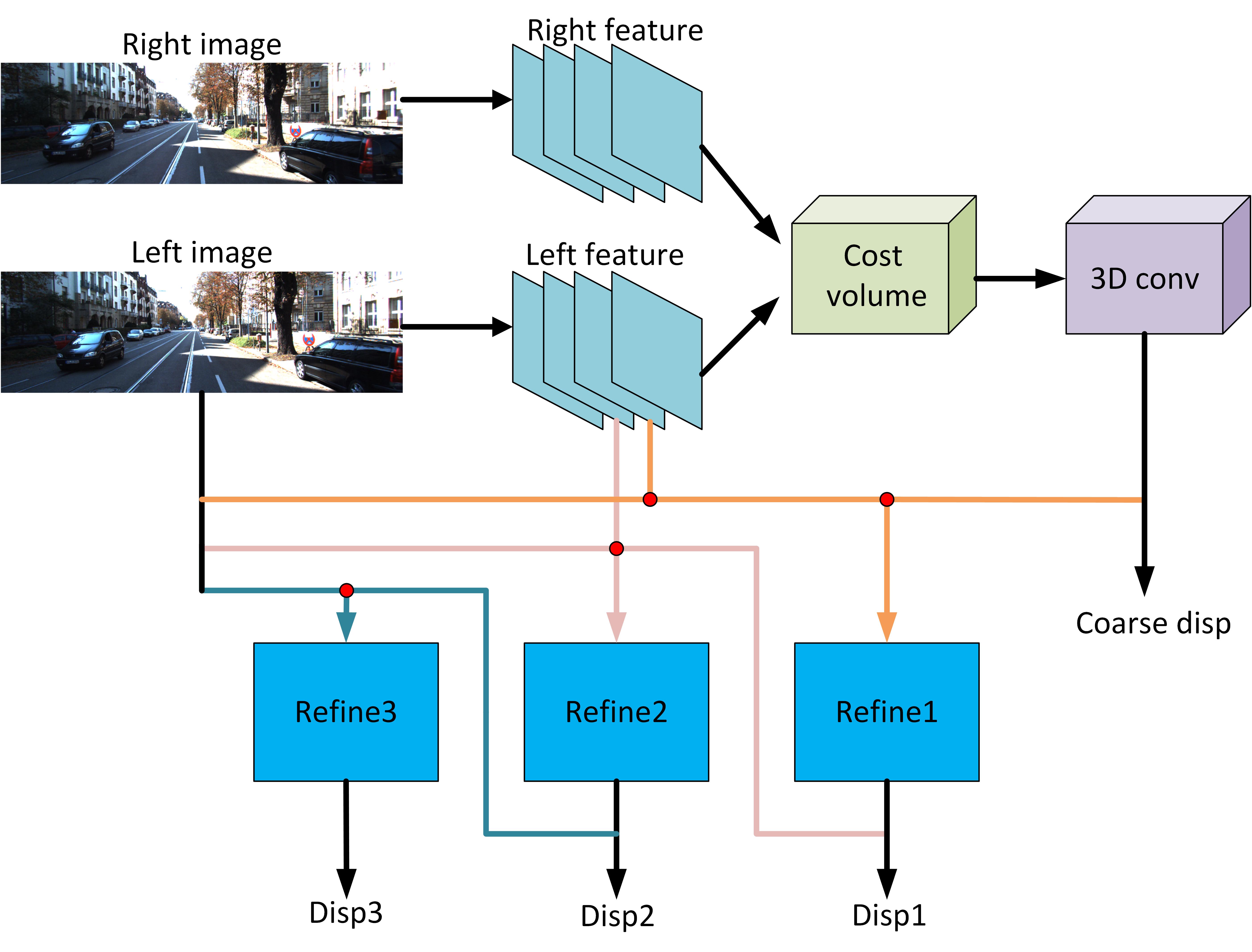}
			\end{minipage}
		\end{center}
		\caption{Our baseline network architecture. The red dot denotes concatenation opreation.}
		\label{fig:network}
	\end{figure*}

	\begin{table}
		\caption{Network architecture.}
		\footnotesize
		\begin{center}
			{\setlength{\tabcolsep}{1.5mm}
				\begin{tabular}{l|cccccc}
					\toprule  
					Layer &  Kernel & Stride & Channels & In & Out & Input \\ 
					\midrule         
					conv1   & $5\times5$ & 2 &  3/32&  1 & 2 & {\tabincell{c}{Stereo- \\ -images}}  \\
					conv2   & $5\times5$ & 2 &  32/32&  2 & 4 & conv1  \\
					conv3   & $5\times5$ & 2 &  32/32&  4 & 8 & conv2  \\
					res1   & $3\times3$ & 1 &  32/32&  8 & 8 & conv3  \\
					res2   & $3\times3$ & 1 &  32/32&  8 & 8 & res1  \\
					res3   & $3\times3$ & 1 &  32/32&  8 & 8 & res2  \\
					res4   & $3\times3$ & 1 &  32/32&  8 & 8 & res3  \\
					res5   & $3\times3$ & 1 &  32/32&  8 & 8 & res4  \\
					res6   & $3\times3$ & 1 &  32/32&  8 & 8 & res5  \\
					conv4   & $3\times3$ & 1 &  32/32&  8 & 8 & res6  \\
					\midrule
					3dconv1  & $3\times3\times3$ & 1 &  32/32&  8 & 8 & cost  \\
					3dBN1  & - & 1 &  32/32&  8 & 8 & 3dconv1  \\
					3dconv2  & $3\times3\times3$ & 1 &  32/32&  8 & 8 & 3dBN1  \\
					3dBN2  & - & 1 &  32/32&  8 & 8 & 3dconv2  \\
					3dconv3  & $3\times3\times3$ & 1 &  32/32&  8 & 8 & 3dBN2  \\
					3dBN3  & - & 1 &  32/32&  8 & 8 & 3dconv3  \\
					3dconv4  & $3\times3\times3$ & 1 &  32/32&  8 & 8 & 3dBN3  \\
					3dBN4  & - & 1 &  32/32&  8 & 8 & 3dconv4  \\
					3dconv5  & $3\times3\times3$ & 1 &  32/1&  8 & 8 & 3dBN4  \\
					\midrule
					billnear1 & - & - &  32/32 &  8 & 4 & coarse1  \\
					concat1 &   - & -  &  -/65&  4 & 4 & {\tabincell{c}{billnear1, \\conv2,   \\ left image}} \\
					atrous1& $3\times3$ & 1 &  65/1&  4 & 4 & concat1 \\
					billnear2 &  - & - &  32/32&  4 & 2 & atrous1  \\
					concat2 &   - & - &  -/65&  2 & 2 & {\tabincell{c}{billnear2, \\conv1,   \\ left image}} \\
					atrous2 & $3\times3$ & 1 &  65/32&  2 & 2 & concat2 \\
					billnear3 & - & - &  32/32&  2 & 1 & atrous2 \\
					concat3 &  - & -  &  -/35&  1 & 1 & {\tabincell{c}{billnear3, \\ left image}} \\
					atrous3 & $3\times3$ & 1 &  65/32&  1 & 1 & concat3 \\
					conv5   & $3\times3$ & 1 &  32/1&  1 &1& atrous3  \\
					relu1  & - & -&  1/1&  1 &1& conv5  \\
					\bottomrule 
			\end{tabular}}
		\end{center}
		\label{Tab:Net}
	\end{table}

	\begin{table}
		\caption{Atrous block.}
		\footnotesize
		\begin{center}
			{\setlength{\tabcolsep}{1.5mm}
				\begin{tabular}{l|ccccc}
					\toprule  
					Layer &Dilate rate & Channels & In & Out & Input \\ 
					\midrule
					conv1  &  - &  65/32&  -& - & input featrues  \\
					atrous res1   & 1 &  32/32&  -& - & conv1 \\
					atrous res2   & 2 &  32/32&  -& - & atrous res1  \\
					atrous res3   & 4 &  32/32&  -& - & atrous res2 \\
					atrous res4   & 8 &  32/32&  -& - & atrous res3 \\
					atrous res5   & 1 &  32/32&  -& - & atrous res4 \\
					atrous res6   & 1 &  32/32&  -& - & atrous res5 \\
					conv2  & - &  32/32&  -& - & atrous res5 \\
					relu1 & - &  32/32&  -& - & conv2 \\
					\bottomrule 
			\end{tabular}}
		\end{center}
		\label{Tab:Atrous}
	\end{table}

	Our proposed occlusion-aware training method can be applied in most stereo methods that make use of reconstruction loss, and the post-processing strategy can be used in any stereo methods with dense disparity output. However, a specific design of network architecture and loss calculation is introduced, for the convenience of explanation and experiments. 
		

	\begin{table*}
			\begin{center}
			\caption{Quantitative results on KITTI  2012 and 2015\cite{KITTI} validation set.}
		\label{stereo_depth}
					\begin{tabular}{l|c|ccccc|ccr}
						\toprule  
						\multirow{2}{*}{Approaches} & \multirow{2}{*}{Dataset} & \multicolumn{5}{c|}{Lower is better} & \multicolumn{3}{c}{Higher is better}\\  [3pt]
						~  & ~ & Abs Rel & Sq Rel & RMSE & RMSE log & D1-all & $\delta<1.25$ & $\delta<1.25^2$ & $\delta<1.25^3$\\ [3pt]
	
						\midrule         
	                   	 OASM-Net  \cite{Li} &\multirow{4}{*}{\tabincell{c}{KITTI \\ 2012}} & \verb|\|    &\verb|\|     & \verb|\|    &\verb|\|     & 6.690\%       & \verb|\|    & \verb|\|    &\verb|\|   \\ [3pt]
						Godard et al\cite{Godard} & ~ & 0.049       & 0.356       & 2.525       & 0.109       & 8.771\%       & 0.959       & 0.985       & 0.993       \\ [3pt]
						{\bf GOUSM(Ours) }                     & ~    &{\bf 0.034}  & {\bf 0.272} & {\bf 2.126} & {\bf 0.089} & {\bf 5.526\%} & {\bf 0.975} & {\bf 0.990} & {\bf 0.995} \\
	
						\midrule         
						Zhou et al.\cite{Zhou}     &  \multirow{8}{*}{\tabincell{c}{KITTI \\ 2015}}      & \verb|\|    & \verb|\|    & \verb|\|    & \verb|\|    & 8.35\%        & \verb|\|    & \verb|\|    & \verb|\|    \\[3pt]
						SegStereo\cite{SegStereo}    & ~      & \verb|\|    & \verb|\|    & \verb|\|    & \verb|\|    & 7.70\%        & \verb|\|    & \verb|\|    & \verb|\|    \\ [3pt]
	                    Luo et al.\cite{Luo}                     & ~    & \verb|\|    &\verb|\|     & \verb|\|    &\verb|\|     & 6.310\%       & \verb|\|    & \verb|\|    &\verb|\|  \\  [3pt]
	                   	 OASM-Net \cite{Li}			    & ~       & \verb|\|    &\verb|\|     & \verb|\|    &\verb|\|     & 6.650\%       & \verb|\|    & \verb|\|    &\verb|\|   \\ [3pt]
	                    Teng et al.\cite{Teng}    			   & ~    & 0.149       & 1.223       & 5.732       & 0.225       & \verb|\|      & 0.829       & 0.944       & 0.978       \\[3pt]
						Godard et al\cite{Godard} & ~ & 0.068       & 0.835       & 4.392       & 0.146       & 9.194\%       & 0.942       & 0.978       & 0.989       \\ [3pt]
						UnOS \cite{UnOS}        & ~   & 0.060       & 0.833       & 4.187       & 0.135       & 7.073\%       & 0.955       & 0.981       & 0.990       \\ [3pt]
						{\bf GOUSM(Ours) }                     & ~    &{\bf 0.054}  & {\bf 0.739} & {\bf 3.898} & {\bf 0.124} & {\bf 5.530\%} & {\bf 0.966} & {\bf 0.985} & {\bf 0.992} \\
						\bottomrule 
				\end{tabular}
			\end{center}
	\end{table*}

As shown in Tab. \ref{Tab:Net}  and Fig. \ref{fig:network}, our network, inspired by StereoNet\cite{StereoNet}, is a U-net like architecture, which contains sub-sampling layers and skip connection layers. Following previous methods, Siamese architecture that each branch processes the left and right images is employed to gain image features and discrimination along the disparity range. Specifically, left and right images as input and are passed through 3 convolution layers with stride 2 for downsampling, and then 6 resnet blocks are used to learn deep features. At the end of them, a convolution layer is appended without any activation. Then, image features of left and right are fed to build cost volume which is as described in \cite{GWC}. The cost volume is passed to 3D convolution blocks that each one contains a 3D convolution layer, batch normalization, and a leaky ReLU activation to exploit the spatial relationship of disparity so that the coarse disparity map is derived. To further refine the disparity, we take advantage of a refinement module similar to \cite{StereoNet}, i.e., task the refinement module of only finding a residual to add or subtract from the coarse prediction. The output is then passed through 3 refine parts. Each part consists of image resizing, feature concat, and an atrous block. Coarse disparity map bilinearly upsampled as well as the original left RGB image resized and the previous feature jointly concat to be formed as the input of atrous block. This block mainly contains 6 residual blocks. Every block employs 3 × 3 atrous convolutions, batch-normalization, and a leaky ReLu activation, and its output is a 1-dimensional disparity residual map that is then added to the previous prediction, Details of the atrous block can be found in Tab. \ref{Tab:Atrous}.


	\begin{figure*}[t]
		\begin{center}
			\begin{minipage}[t]{1\linewidth}
				\centering
				\includegraphics[width=1\linewidth]{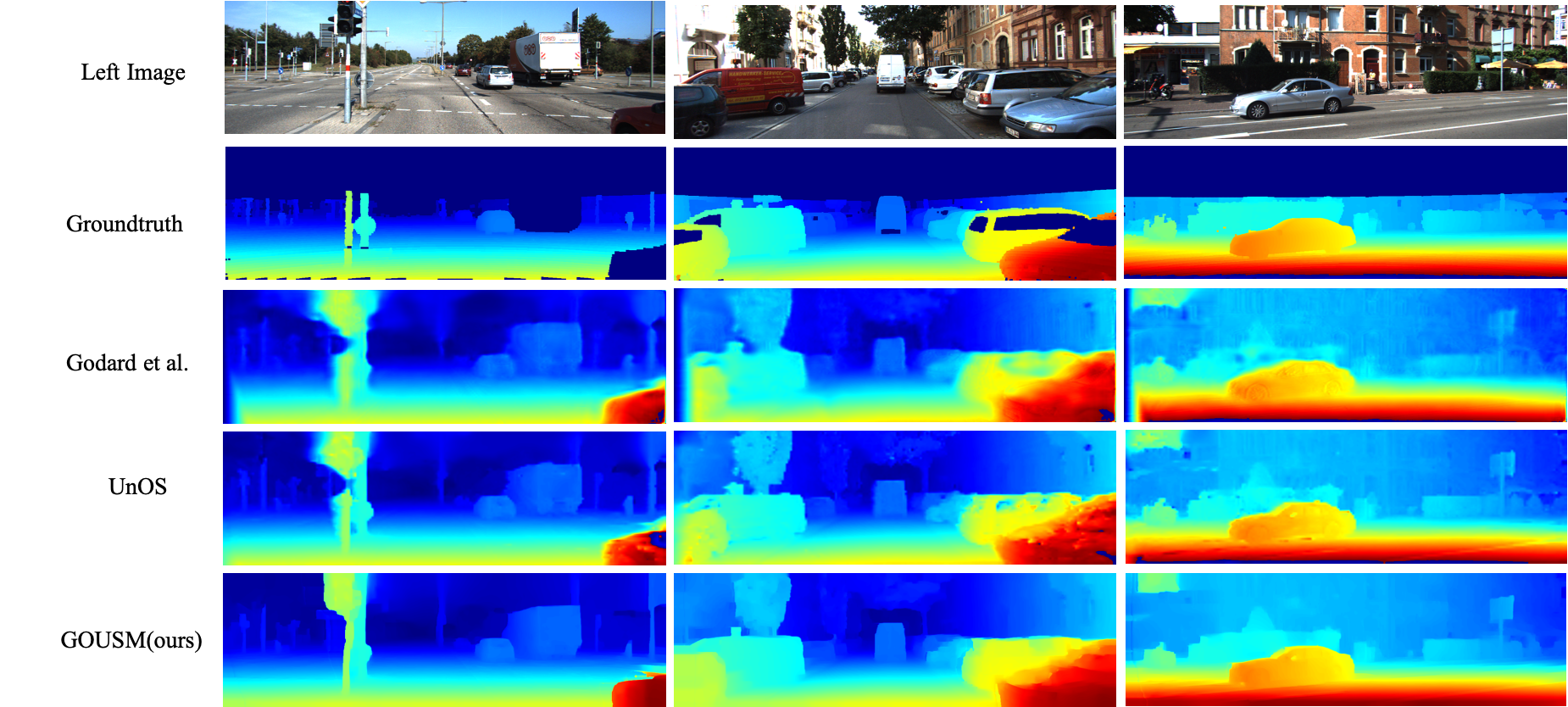}
			\end{minipage}
		\end{center}
		\caption{Qualitative comparison of different methods on KITTI 2015. Ours shows the smoothest results and gives the cleanest predictions around edges, where occluded regions usually exist.}
		\label{fig:KITTI-comparision}
		
		\label{fig:onecol}
	\end{figure*}

	\subsection{Loss Calculation}
	A weighted sum of reconstruction loss and smooth loss is calculated for every pixel. Then, the occlusion mask is applied to exclude the occluded pixels and the final loss is the average value of all kept pixel-wise losses.  
		
	\subsubsection{Reconstruction Loss}
	
	The left image is reconstructed by sampling from the right image according to predicted disparities, the same as the method used in \cite{Godard}. Structure Similarity loss (SSIM)\cite{SSIM} and L1 loss are used to measure the reconstruction quality, 
	\begin{equation}
	C_{i,j}^{r} = \alpha \frac{1\!\!-\!\!SSIM(I_{i,j}^l\!, \widetilde{I}_{i,j}^l\!)}{2}\!+\!(1\!-\!\alpha)\!\!\left\|\!(I_{i,j}^l\!- \widetilde{I}_{i,j}^l)\!\right\|\!, 
	\end{equation}
	where $I_{i,j}^l$ and $\widetilde{I}_{i,j}^l$ stand for the intensity values on the original and recovered left image at location $(i, j)$, respectively; $\alpha$ is the weight and $0 \le \alpha \le 1$.
	
	The final pixel-wise reconstruction loss is an aggregation in different directions using Adaptive Support Weights(ASW) proposed in \cite{ASN}. Cost $C_{i, j}^{ar}$ aggregated in a window of size $2k\times2k$ is:
	\begin{equation}
	C_{i, j}^{ar}=\frac{\sum_{x=i-k}^{i+k-1}\sum_{y=j-k}^{j+k-1}w_{x,y}C_{x,y}^{r}}{\sum_{x=i-k}^{i+k-1}\sum_{y=j-k}^{j+k-1}w_{x,y}},
	\end{equation}
	where $w_{x,y}=e^{-\frac{I_{i,j}-I_{x,y}}{2}}$ and $I_{i,j}$ denotes intensity of pixel$(i, j)$ in input left image. 
	
	\subsubsection{Smooth Loss}
	Edge-aware smooth loss\cite{Godard} is used as a regularization term, to make the prediction locally smooth. 
	The smooth loss on pixel $(i,j)$ is 
	\begin{equation}
	C_{i, j}^s = \frac{1}{N}\!\sum\left|\partial_xd_{i,j}^l\right|e^{-\left\|\partial_xI_{i,j}^l\right\|}\! + \!\left|\partial_yd_{i,j}^l\right|e^{-\left\|\partial_yI_{i,j}^l\right\|}
	\end{equation}
	where $\partial d$, $\partial I$ denote the gradients of disparity and intensity, respectively.
	\subsubsection{Occlusion-Aware Loss}
	The total loss on pixel $(i,j)$ is 
	\begin{equation}
	C_{i,j} = w_1C_{i,j}^{ar}+w_2C_{i,j}^{s}, 
	\end{equation}
	where $w_1, w_2$ are the weights.
	
	The overall loss on the whole image is 
	\begin{equation}
	C = \frac{1}{\sum_{i,j}M(i,j)}\sum_{i,j} M(i, j) C_{i,j},
	\end{equation}
	where $M$ is the occlusion mask, where values for occluded pixels are zeros. 
	
	In this way, occlusions are excluded in loss calculation.

	\section{Experiments}
	
	\begin{table*}
		\label{ablation}
		\caption{Ablation study. Neither GOAT nor GOAPP is used in the baseline.}
		\begin{center}
				\begin{tabular}{l|ccccc|ccr}
					\toprule  
					\multirow{2}{*}{Approaches} & \multicolumn{5}{c|}{Lower is better} & \multicolumn{3}{c}{Higher is better}\\  
					~  & Abs Rel & Sq Rel & RMSE & RMSE log & D1-all & $\delta<1.25$ & $\delta<1.25^2$ & $\delta<1.25^3$\\ 
					\midrule         
					Baseline                  & 0.059       & 0.857       & 4.132       & 0.135       & 6.834\%       & 0.958       & 0.980       & 0.989       \\ [3pt]
					Baseline+GOAPP             & 0.059       & 0.673       & 3.902       & 0.128       & 6.410\%       & 0.960       & 0.983       & 0.991       \\[3pt]
					Baseline+GOAT              & 0.056       & 0.768       & 4.032       & 0.131       & 6.047\%       & 0.962       & 0.982       & 0.990       \\[3pt]
					 \tabincell{c}{GOUSM(Baseline+ \\ GOAT+GOAPP)}        & {\bf 0.054} & {\bf 0.739} & {\bf 3.898} & {\bf 0.124} & {\bf 5.530\%} & {\bf 0.966} & {\bf 0.985} & {\bf 0.992} \\
					\bottomrule 
			\end{tabular}
		\end{center}
		\label{ablation}
	\end{table*}
	
			\begin{table}
		\caption{Quantitative results on the FlyingThings3D test set.}
		\begin{center}
			{\setlength{\tabcolsep}{1.5mm}
				\begin{tabular}{l|ccc}
					\toprule  
					Approaches & Supervised & EPE & 3px-error\\ 
					\midrule         
					PSMNet\cite{PSMNet}   & Yes & 1.09 & \verb|\|   \\[3pt]
					GC-Net Full\cite{GC-Net}   & Yes & 2.51 & 9.3\%  \\ [3pt]
					GC-Net Fast\cite{GC-Net}   & Yes    &7.27 &      24.2\%    \\
\midrule 
		SGM\cite{SGM}   & No   & 8.70 &      \verb|\|  \\[3pt]
		Godard et al.\cite{Godard}   & No    & 8.28 &      28.7\%  \\[3pt]
		{\bf GOUSM(Ours)}   &No    &  {\bf 6.22} &       {\bf 16.2\%}   \\

					\bottomrule 
			\end{tabular}}
		\end{center}
		\label{Tab:Quantitative-results-Flying}
	\end{table}

	\begin{table*}
		\label{tab-Godard-post-processing}
		\caption{Quantitative results of GOAPP extended to Godard et al.\cite{Godard} on the KITTI 2015 validation set. 'All' in the table represents the measured area from KITTI, where 'No' refers to regions for which the matching correspondence is inside the image domain,  'Yes' refers to all image regions for which groundtruth could be measured, including regions which map to points outside the image domain in the other view.}
		\begin{center}
				\begin{tabular}{l|c|ccccc|cccr}
					\toprule  
					\multirow{2}{*}{Approaches} & \multirow{2}{*}{All} & \multicolumn{5}{c|}{Lower is better} & \multicolumn{3}{c}{Higher is better}\\  
					~ & ~ & Abs Rel & Sq Rel & RMSE & RMSE log & D1-all & $\delta<1.25$ & $\delta<1.25^2$ & $\delta<1.25^3$\\ 
					\midrule         
					Godard et al.              & No  & {\bf 0.068} & {\bf 0.835} & 4.392       & 0.146       & 9.194\%        & 0.942       & 0.978       & 0.989 \\[3pt]
					 \tabincell{c}{Godard et al.+ \\ GOAPP }       & No  & {\bf 0.068} & 0.914       &{\bf  4.364} & {\bf 0.140} & {\bf 8.618\%}  & {\bf 0.947} & {\bf 0.981} & {\bf 0.990} \\[3pt]
					\cmidrule{1-10}
					Godard et al.              & Yes & 0.102       & 2.280       & 5.728       & 0.202       & 10.755\%       & 0.928       & 0.966       & 0.980 \\[3pt]
					\tabincell{c}{Godard et al.+ \\ GOAPP }         & Yes & {\bf 0.069} & {\bf 0.923} & {\bf 4.368} & {\bf 0.142} & {\bf 8.956\%}  & {\bf 0.946} & {\bf 0.980} & {\bf 0.990} \\
					\bottomrule 
			\end{tabular}
		\end{center}
		\label{tab-Godard-post-processing}
	\end{table*}
	
		\begin{figure*}[t]
		\begin{center}
			\begin{minipage}[t]{1\linewidth}
				\centering
				\includegraphics[width=1\linewidth]{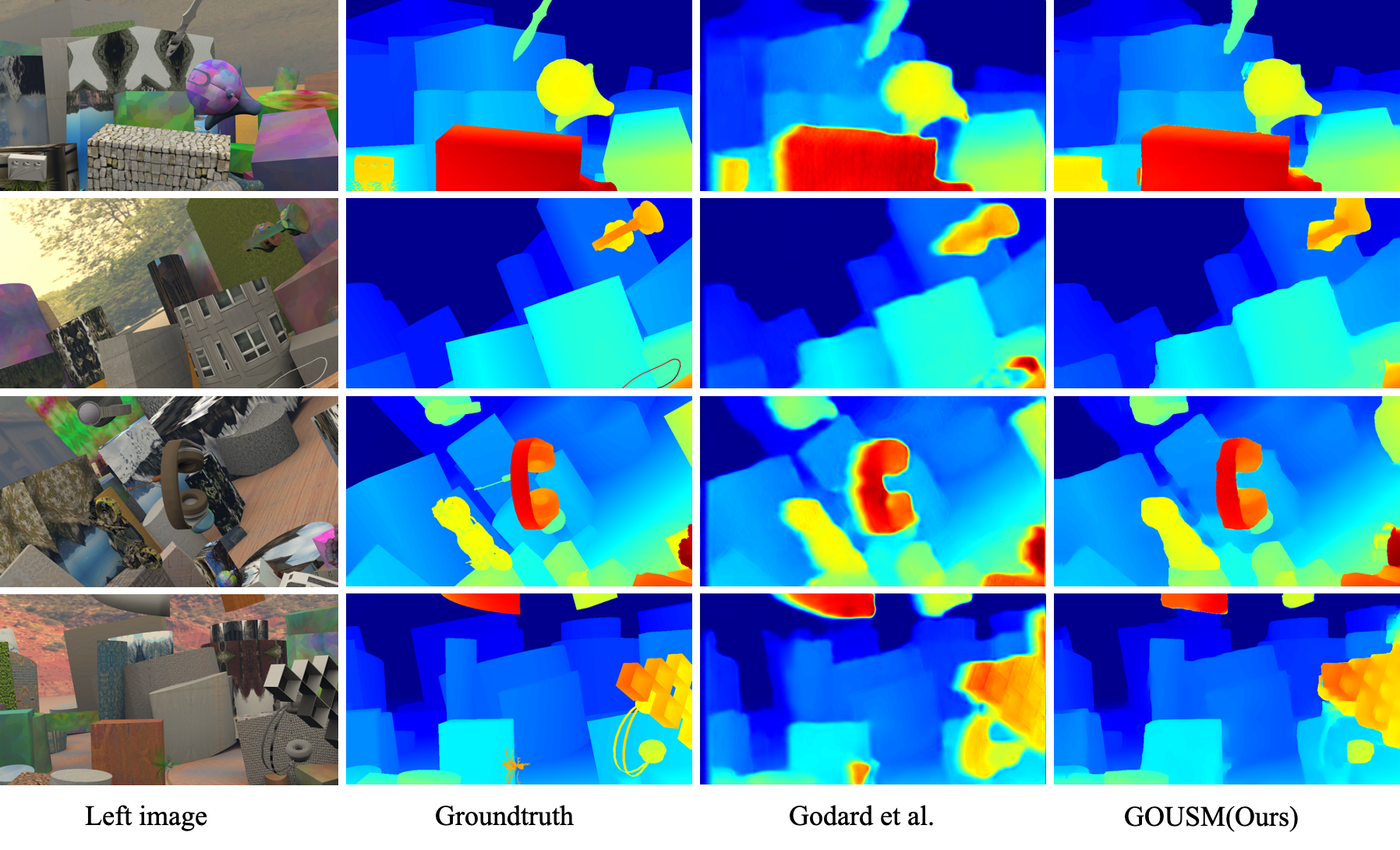}
			\end{minipage}
			
		\end{center}
		\caption{Qualitative comparison of different methods on FlyingThings3D. Best viewed in color.}
\label{Fig:Qualitative-comparison}
	\end{figure*}


	\subsection{Dataset and Evaluation Metrics}
   KITTI\cite{KITTI} raw data covers 61 scenes, containing 42,382 pairs of rectified stereo images without disparity groundtruth. KITTI 2015\cite{KITTI_2015} and KITTI 2012 is the most widely used dataset for supervised stereo matching. It provides 200 and 194 pairs of stereo images with high-quality disparity groundtruth, respectively. Furthermore, some  experiments have been conducted on SceneFlow\cite{DispNet}, in which FlyingThings3D is its largest subset, with  21, 818 and 4,248 frames in its training and test set respectively, and Cityscapes\cite{Cityscapes} is another popular dataset for autonomous driving that contains many scenes in the city.

   Standard depth metrics from \cite{Depth_metric} and disparity metric D1-all from \cite{KITTI} are used to measure and compare performances of different methods. For depth metrics, maximum predictions from models are capped to 80 meters, to keep pace with the other literatures\cite{Godard,UnOS,Zhou,SegStereo}. Depth groundtruth is derived from disparity, and the error calculated in depth space is more sensitive for small disparity values since disparity is in inverse proportion to depth.

Some commonly used depth metrics are listed as follows:
\begin{itemize}
	\item Threshold: $\% \hspace{3pt} of \hspace{3pt} y_i \hspace{3pt} s.t.\hspace{3pt} max(\frac{y_i}{y_i^*}, \frac{y_i^*}{y_i})=\delta < thr$
	\item Absolute Relative Difference (Abs Rel): $\frac{1}{\left|T\right|}\sum_{y\in{T}}\left|y-y^*\right|/y^*$, 
	\item Squared Relative Difference (Sq Rel), :$\frac{1}{\left|T\right|}\sum_{y\in{T}}\left|\left|y-y^*\right|\right|^2/y^*$
	\item Root of Mean Squre Error(RMSE), $\sqrt{\frac{1}{\left|T\right|}\sum_{y\in{T}}\left|\left|y_i-y_i^*\right|\right|^2}$, 
   \item RMSE(log): $\sqrt{\frac{1}{\left|T\right|}\sum_{y\in{T}}\left|\left|logy_i-logy_i^*\right|\right|^2}$
\end{itemize}

	\subsection{Implementation Details}
	The whole pipeline is implemented in Python and Tensorflow\cite{Tensorflow}.  RMSProp\cite{RMSProp} is used for optimization with an exponentially-decaying learning rate initially set to $0.001$.  The batch size is set to 1. In loss calculation, the values of $w_1$, $w_2$, $\alpha$, $k$ are set to 0.85, 0.15, 0.8, 16, respectively. Following other unsupervised methods, our model is trained on the remaining 30k stereo samples from KITTI raw data excluding scenes of KITTI 2015, for about 20 epochs. The whole training process takes about 2 days on an Nvidia 1080Ti GPU.

	\subsection{Results on KITTI}
	As is shown in Tab. \ref{stereo_depth}, our occlusion-aware approach performs better than the other state-of-the-art unsupervised methods, proving the effectiveness of our method. Most methods in the table deal with occlusion using left-right consistency check, i.e.,\cite{Godard,Zhou,SegStereo,UnOS,Luo} , or direct CNN prediction, i.e., OASM-Net\cite{Li}. Fig. \ref{fig:KITTI-comparision} compares the disparity outputs qualitatively in KITTI 2015 validation set. Compared with their outputs, the disparities from our method tend to have more smooth and clearer near edges, where occluded regions usually exist.

		\begin{figure*}[htbp]
		\begin{center}
			\begin{minipage}[t]{0.9\linewidth}
				\centering
				\includegraphics[width=1\linewidth]{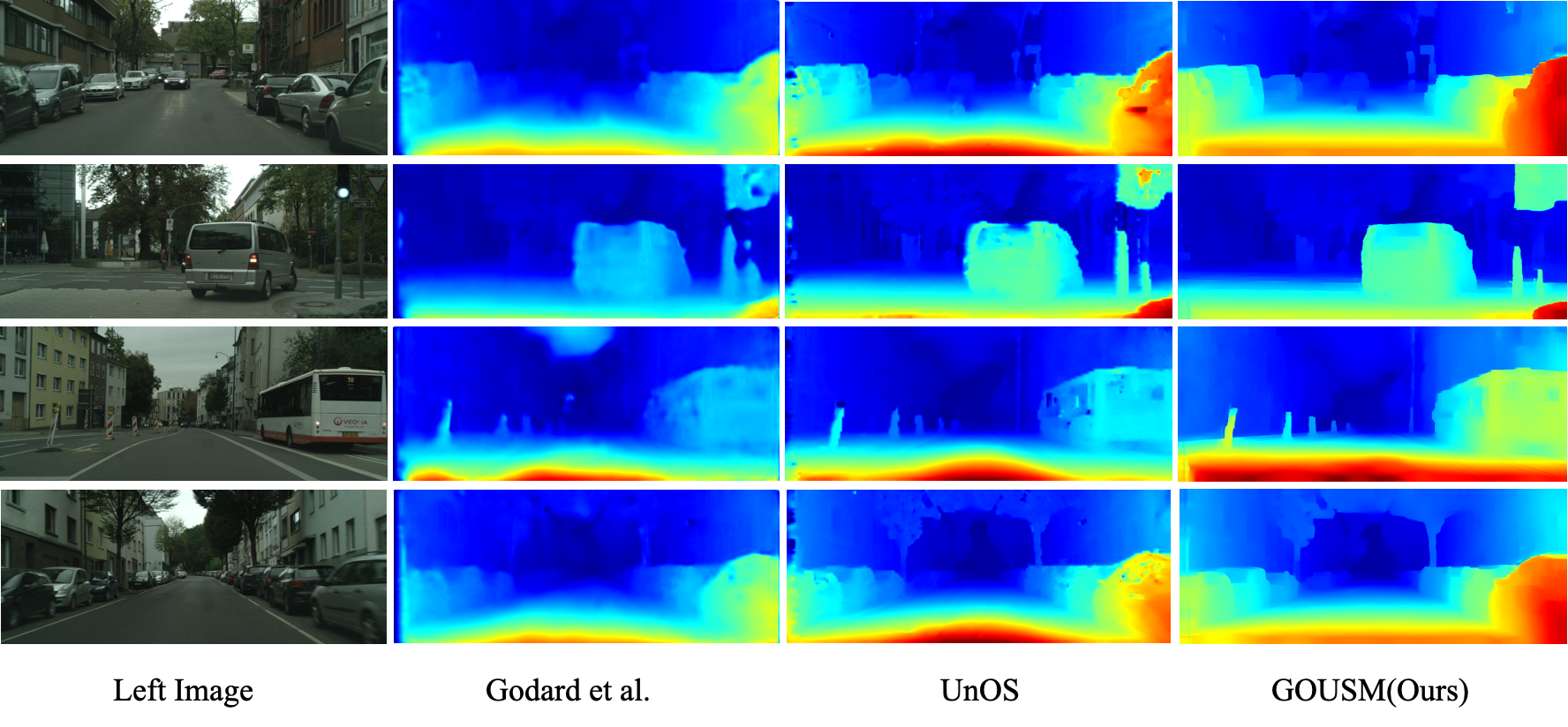}
			\end{minipage}
		\end{center}
		\caption{Qualitative results in Cityscapes. Our approach outperforms other methods. Even for ill-posed regions such as glass, we still get promising results and preserve many details. Best viewed in color.}
		\label{fig:extend-CityScapes}
	\end{figure*}

			\begin{figure*}[htbp]
		\begin{center}
			\begin{minipage}[t]{1\linewidth}
				\includegraphics[width=1\linewidth]{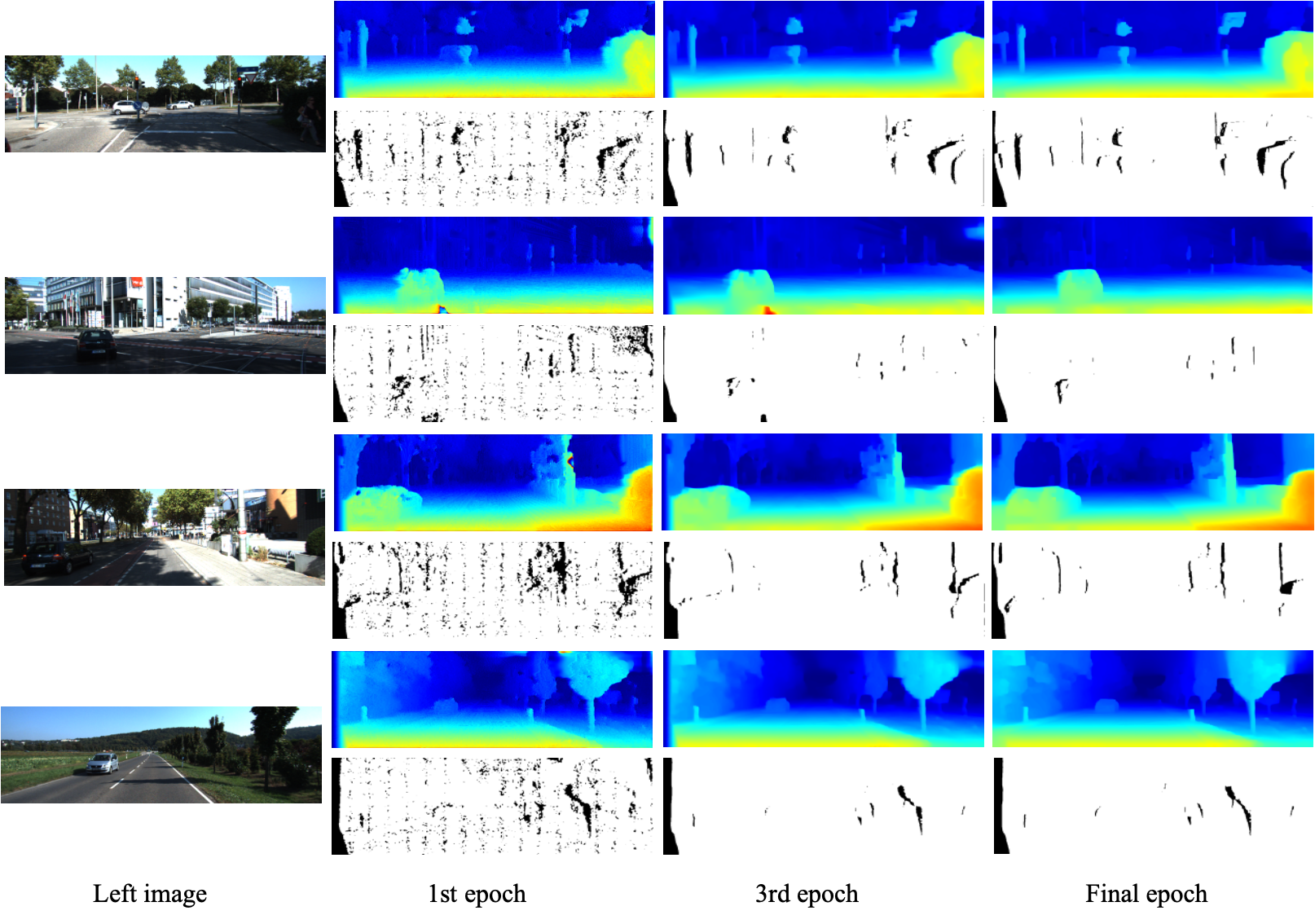}
			\end{minipage}
		\end{center}
		\caption{Disparity maps and occlusion masks during training from different epochs. Black in occlusion masks refers to occlusions. As the training goes on, the qualities of both disparities and occlusions increase gradually. Best viewed in color.}
\label{iter}
	\end{figure*}

\subsection{Results on SceneFlow}
	Fig.  \ref{Fig:Qualitative-comparison} and Tab. \ref{Tab:Quantitative-results-Flying} shows the results on SceneFlow. Since FlyingThings3D is a large-scale synthesized dataset, and all samples are generated with precise dense groundtruths, the state-of-the-art supervised methods behave much better than unsupervised methods. However, the yield of such full and high-quality dense groundtruth is usually impossible in real-world scenarios. Most stereo data from the real-world is born without groundtruth and is suitable for unsupervised methods only. Our method performs the best among all compared unsupervised methods.

	\begin{figure}[t]
		\begin{center}
			\begin{minipage}[t]{1\linewidth}
				\centering
				\includegraphics[width=1\linewidth]{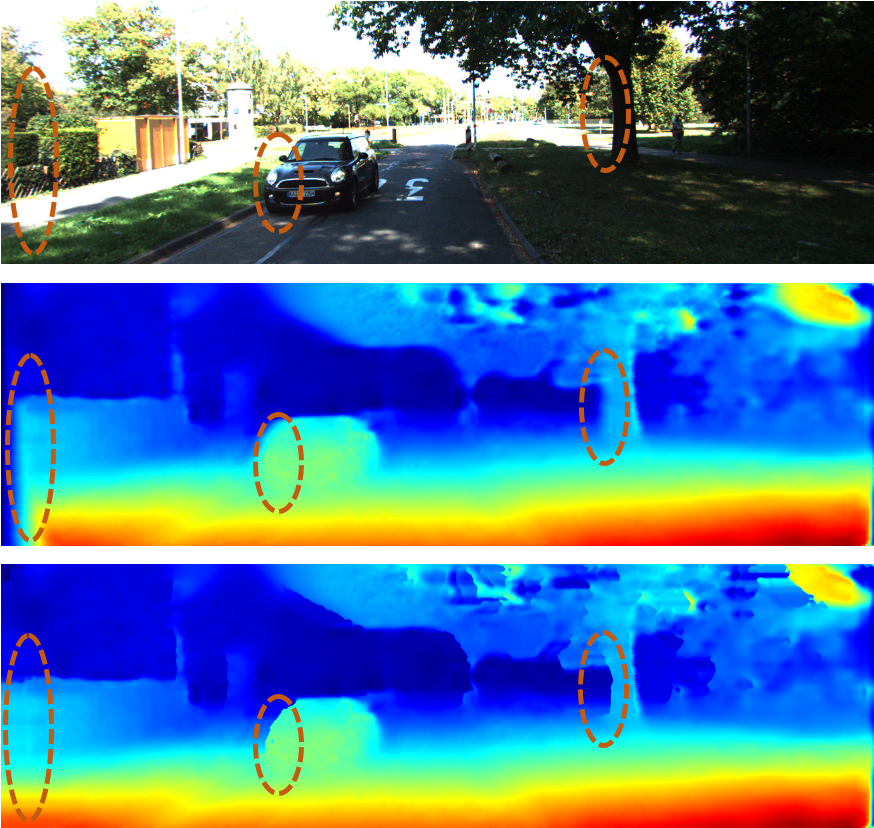}
			\end{minipage}
		\end{center}
		\caption{Qualitative results of extending GOAPP to Godard et al.\cite{Godard}.  Occlusion regions tend to be blurred in the original method, and become clear after GOAPP. From top to bottom: left image, Godard et al., Godard et al.+GOAPP. Best viewed in color.}
		\label{fig:extend-post-processing}
		\label{fig:onecol}
	\end{figure}

	\subsection{Ablation Experiments}
	To show the contribution of being occlusion-aware during training and post-processing, some ablation experiments are conducted, and the results are shown in Tab. \ref{ablation}. The adoption of GOAT contributes to an improvement of about 0.8\% in D1-all, the performance already a litter better than the other state-of-the-art methods shown in Tab. \ref{stereo_depth}, and GOAT + GOAPP gives an improvement of 1.3\%. It means the relative error w.r.t. D1-all has been reduced by 11.7\%, 19.1\%, respectively.

	\subsection{Extending to the other methods}
	As has been mentioned in Sec.4.1, our proposed occlusion-aware pipeline can be easily extended to the other stereo matching methods, by adopting the GOAT or GOAPP, or both of them. Two experiments as application examples have been conducted to demonstrate the advantages of occlusion-aware strategies in stereo matching.

 Dense disparity prediction is the only requirement for the adoption of GOAPP. Tab. \ref{tab-Godard-post-processing} and Fig. \ref{fig:extend-post-processing} show the results of Godard's\cite{Godard} method after adding GOAPP as the last post-processing step. When only considering the results from common regions, GOAPP brings improvements for most metrics. After taking all regions into consideration, including the exclusive regions on left images, the improvements are quite obvious. It means that GOAPP does work in removing outputs of low-quality in post-processing.

	\begin{table}
		\label{tab-supervised-combined}
		\caption{Quantitative results of extending GOAT to supervised stereo methods, by providing pre-trained models. KITTI 2015 dataset is randomly split into training and validation subset, containing 160 and 40 samples, respectively. PSMNet\cite{PSMNet} and GwcNet\cite{GWC} are trained with their public code using our batch size.}
		\begin{center}
		\footnotesize
			\begin{tabular}{ll|ccc}
				\toprule  
				\centering
				Approaches & {\tabincell{c}{Training \\ mode}}  & D1-all &  EPE (pixels) & Times (ms)\\ 
				\midrule         
				\multirow{2}{*}{StereoNet\cite{StereoNet}} & Supervised   &  5.827\%         &  1.214       & \multirow{2}{*}{85}  \\
				~                                          & Combined     &  {\bf 4.058\%}   & {\bf 0.946}  & ~  \\
				\midrule 
				\multirow{2}{*}{PSMNet\cite{PSMNet}}       & Supervised   &  3.454\%         & 0.918        &  \multirow{2}{*}{410}  \\ 
				~                                          & Combined     &  {\bf 3.193\%}   &  {\bf 0.854} & ~  \\
				\midrule 
				\multirow{2}{*}{GwcNet\cite{GWC}}          & Supervised   &  2.825\%         &  0.821       &  \multirow{2}{*}{330}  \\
				~                                          & Combined     &  {\bf 2.651\%}   & {\bf 0.767}  & ~ \\
				\bottomrule 
			\end{tabular}
		\end{center}
		\label{tab-supervised-combined}
	\end{table}

 Most supervised stereo methods need pretraining on the synthetic dataset SceneFlow\cite{DispNet} because the size of an available real-scene dataset with high-quality groundtruths is far from large enough. For example, KITTI provides only hundreds of data samples with high groundtruth, and tens of thousands of samples in its raw data, however, without groundtruth. And therefore, GOAT can provide help to supervised methods by providing high-quality pre-trained models and avoiding the domain shift from pretraining to finetuning. Tab. \ref{tab-supervised-combined} shows the results of such combinations. All the listed methods yield improvements.
 
	\subsection{ Generalize to other datasets}
  In this section, we show the generalization ability of our approach and illustrate some examples from Cityscapes. Our model is trained on KITTI and conducts inferences on images from Cityscapes\cite{Cityscapes} without finetuning. Some results are shown in Fig. \ref{fig:extend-CityScapes}. Despite the fact that our model has not been trained on Cityscapes, our method still shows good results and performs better than the other approaches. It's a strong implication of the generalization ability and robustness of our approach, even for the scenes never seen, or image data from different cameras with different parameters.

\subsection{The gradual improvement of disparity and occlusion mask}
As shown in Fig.  \ref{iter}, in early epochs, the quality of disparities tends to be poor, so as the resulting occlusion masks. However, their qualities keep improving as the training goes on. This process implies that our proposed method, i.e., detecting occlusion from predicted disparity directly and iteratively, is working well.

	\section{Conclusion}
	In this paper, we point out that occlusion imposes negative influences on unsupervised stereo matching methods because the reconstruction losses from occluded regions are meaningless and harmful for model training. A novel geometry-based method is proposed to detect occluded regions, resulting in an occlusion mask. It can then be used in training and post-processing iteratively to deal with occlusions. A specific design of network architecture and loss calculation is introduced for the convenience of explanation and experiments. It is then shown by experiments that this occlusion-aware design could achieve better performances than the other unsupervised methods, without or with different strategies of occlusion handling. What's more, it's worth noting that the proposed occlusion-aware strategies, i.e., GOAT and GOAPP, can be easily extended to the other stereo methods and achieve performance gains. Three extending approaches have been experimented in this paper. However, many other possible ways to make use of Occlusion-Awareness can be explored in the future.

\ifCLASSOPTIONcaptionsoff
  \newpage
\fi



%

%
%

\bibliography{IEEEtran}

\begin{thebibliography}{10}
\providecommand{\url}[1]{#1}
\csname url@samestyle\endcsname
\providecommand{\newblock}{\relax}
\providecommand{\bibinfo}[2]{#2}
\providecommand{\BIBentrySTDinterwordspacing}{\spaceskip=0pt\relax}
\providecommand{\BIBentryALTinterwordstretchfactor}{4}
\providecommand{\BIBentryALTinterwordspacing}{\spaceskip=\fontdimen2\font plus
\BIBentryALTinterwordstretchfactor\fontdimen3\font minus
  \fontdimen4\font\relax}
\providecommand{\BIBforeignlanguage}[2]{{%
\expandafter\ifx\csname l@#1\endcsname\relax
\typeout{** WARNING: IEEEtran.bst: No hyphenation pattern has been}%
\typeout{** loaded for the language `#1'. Using the pattern for}%
\typeout{** the default language instead.}%
\else
\language=\csname l@#1\endcsname
\fi
#2}}
\providecommand{\BIBdecl}{\relax}
\BIBdecl

\bibitem{Godard}
C.~Godard, O.~Mac~Aodha, and G.~J. Brostow, ``Unsupervised monocular depth
  estimation with left-right consistency,'' in \emph{Proceedings of the IEEE
  Conference on Computer Vision and Pattern Recognition}, 2017, pp. 270--279.

\bibitem{Zhou}
C.~Zhou, H.~Zhang, X.~Shen, and J.~Jia, ``Unsupervised learning of stereo
  matching,'' in \emph{Proceedings of the IEEE International Conference on
  Computer Vision}, 2017, pp. 1567--1575.

\bibitem{UnOS}
Y.~Wang, P.~Wang, Z.~Yang, C.~Luo, Y.~Yang, and W.~Xu, ``Unos: Unified
  unsupervised optical-flow and stereo-depth estimation by watching videos,''
  in \emph{Proceedings of the IEEE Conference on Computer Vision and Pattern
  Recognition}, 2019, pp. 8071--8081.

\bibitem{SSIM}
Z.~Wang, A.~C. Bovik, H.~R. Sheikh, E.~P. Simoncelli \emph{et~al.}, ``Image
  quality assessment: from error visibility to structural similarity,''
  \emph{IEEE transactions on image processing}, vol.~13, no.~4, pp. 600--612,
  2004.

\bibitem{STN}
M.~Jaderberg, K.~Simonyan, A.~Zisserman \emph{et~al.}, ``Spatial transformer
  networks,'' in \emph{Advances in neural information processing systems},
  2015, pp. 2017--2025.

\bibitem{KITTI_2015}
M.~Menze and A.~Geiger, ``Object scene flow for autonomous vehicles,'' in
  \emph{Proceedings of the IEEE Conference on Computer Vision and Pattern
  Recognition}, 2015, pp. 3061--3070.

\bibitem{SegStereo}
G.~Yang, H.~Zhao, J.~Shi, Z.~Deng, and J.~Jia, ``Segstereo: Exploiting semantic
  information for disparity estimation,'' in \emph{Proceedings of the European
  Conference on Computer Vision (ECCV)}, 2018, pp. 636--651.

\bibitem{Li}
A.~Li and Z.~Yuan, ``Occlusion aware stereo matching via cooperative
  unsupervised learning,'' in \emph{Asian Conference on Computer Vision}.\hskip
  1em plus 0.5em minus 0.4em\relax Springer, 2018, pp. 197--213.

\bibitem{SGM}
H.~Hirschmuller, ``Stereo processing by semiglobal matching and mutual
  information,'' \emph{IEEE Transactions on pattern analysis and machine
  intelligence}, vol.~30, no.~2, pp. 328--341, 2007.

\bibitem{C1}
J.~Zbontar, Y.~LeCun \emph{et~al.}, ``Stereo matching by training a
  convolutional neural network to compare image patches.'' \emph{Journal of
  Machine Learning Research}, vol.~17, no. 1-32, p.~2, 2016.

\bibitem{C2}
W.~Luo, A.~G. Schwing, and R.~Urtasun, ``Efficient deep learning for stereo
  matching,'' in \emph{Proceedings of the IEEE Conference on Computer Vision
  and Pattern Recognition}, 2016, pp. 5695--5703.

\bibitem{C3}
A.~Shaked and L.~Wolf, ``Improved stereo matching with constant highway
  networks and reflective confidence learning,'' in \emph{Proceedings of the
  IEEE Conference on Computer Vision and Pattern Recognition}, 2017, pp.
  4641--4650.

\bibitem{DispNet}
N.~Mayer, E.~Ilg, P.~Hausser, P.~Fischer, D.~Cremers, A.~Dosovitskiy, and
  T.~Brox, ``A large dataset to train convolutional networks for disparity,
  optical flow, and scene flow estimation,'' in \emph{Proceedings of the IEEE
  Conference on Computer Vision and Pattern Recognition}, 2016, pp. 4040--4048.

\bibitem{FCN}
J.~Long, E.~Shelhamer, and T.~Darrell, ``Fully convolutional networks for
  semantic segmentation,'' in \emph{Proceedings of the IEEE conference on
  computer vision and pattern recognition}, 2015, pp. 3431--3440.

\bibitem{PSMNet}
J.-R. Chang and Y.-S. Chen, ``Pyramid stereo matching network,'' in
  \emph{Proceedings of the IEEE Conference on Computer Vision and Pattern
  Recognition}, 2018, pp. 5410--5418.

\bibitem{StereoNet}
S.~Khamis, S.~Fanello, C.~Rhemann, A.~Kowdle, J.~Valentin, and S.~Izadi,
  ``Stereonet: Guided hierarchical refinement for real-time edge-aware depth
  prediction,'' in \emph{Proceedings of the European Conference on Computer
  Vision (ECCV)}, 2018, pp. 573--590.

\bibitem{Corr2}
L.~Yu, Y.~Wang, Y.~Wu, and Y.~Jia, ``Deep stereo matching with explicit cost
  aggregation sub-architecture,'' in \emph{Thirty-Second AAAI Conference on
  Artificial Intelligence}, 2018.

\bibitem{GC-Net}
A.~Kendall, H.~Martirosyan, S.~Dasgupta, P.~Henry, R.~Kennedy, A.~Bachrach, and
  A.~Bry, ``End-to-end learning of geometry and context for deep stereo
  regression,'' in \emph{Proceedings of the IEEE International Conference on
  Computer Vision}, 2017, pp. 66--75.

\bibitem{Corr3}
X.~Song, X.~Zhao, H.~Hu, and L.~Fang, ``Edgestereo: A context integrated
  residual pyramid network for stereo matching,'' in \emph{Asian Conference on
  Computer Vision}.\hskip 1em plus 0.5em minus 0.4em\relax Springer, 2018, pp.
  20--35.

\bibitem{Corr1}
Z.~Liang, Y.~Feng, Y.~Guo, H.~Liu, W.~Chen, L.~Qiao, L.~Zhou, and J.~Zhang,
  ``Learning for disparity estimation through feature constancy,'' in
  \emph{Proceedings of the IEEE Conference on Computer Vision and Pattern
  Recognition}, 2018, pp. 2811--2820.

\bibitem{CRL}
J.~Pang, W.~Sun, J.~S. Ren, C.~Yang, and Q.~Yan, ``Cascade residual learning: A
  two-stage convolutional neural network for stereo matching,'' in
  \emph{Proceedings of the IEEE International Conference on Computer Vision},
  2017, pp. 887--895.

\bibitem{GWC}
X.~Guo, K.~Yang, W.~Yang, X.~Wang, and H.~Li, ``Group-wise correlation stereo
  network,'' in \emph{Proceedings of the IEEE Conference on Computer Vision and
  Pattern Recognition}, 2019, pp. 3273--3282.

\bibitem{MADNet}
A.~Tonioni, F.~Tosi, M.~Poggi, S.~Mattoccia, and L.~D. Stefano, ``Real-time
  self-adaptive deep stereo,'' in \emph{Proceedings of the IEEE Conference on
  Computer Vision and Pattern Recognition}, 2019, pp. 195--204.

\bibitem{Fua}
P.~Fua, ``A parallel stereo algorithm that produces dense depth maps and
  preserves image features,'' \emph{Machine vision and applications}, vol.~6,
  no.~1, pp. 35--49, 1993.

\bibitem{Sun}
J.~Sun, Y.~Li, S.~B. Kang, and H.-Y. Shum, ``Symmetric stereo matching for
  occlusion handling,'' in \emph{2005 IEEE Computer Society Conference on
  Computer Vision and Pattern Recognition (CVPR'05)}, vol.~2.\hskip 1em plus
  0.5em minus 0.4em\relax IEEE, 2005, pp. 399--406.

\bibitem{Teng}
Q.~Teng, Y.~Chen, and C.~Huang, ``Occlusion-aware unsupervised learning of
  monocular depth, optical flow and camera pose with geometric constraints,''
  \emph{Future Internet}, vol.~10, no.~10, p.~92, 2018.

\bibitem{Wang}
Y.~Wang, Y.~Yang, Z.~Yang, L.~Zhao, P.~Wang, and W.~Xu, ``Occlusion aware
  unsupervised learning of optical flow,'' in \emph{Proceedings of the IEEE
  Conference on Computer Vision and Pattern Recognition}, 2018, pp. 4884--4893.

\bibitem{C}
C.~L. Zitnick and T.~Kanade, ``A cooperative algorithm for stereo matching and
  occlusion detection,'' \emph{IEEE Transactions on pattern analysis and
  machine intelligence}, vol.~22, no.~7, pp. 675--684, 2000.

\bibitem{Luo}
N.~Luo, C.~Yang, W.~Sun, and B.~Song, ``Unsupervised stereo matching with
  occlusion-aware loss,'' in \emph{Pacific Rim International Conference on
  Artificial Intelligence}.\hskip 1em plus 0.5em minus 0.4em\relax Springer,
  2018, pp. 746--758.

\bibitem{Rectified}
R.~Hartley and A.~Zisserman, \emph{Multiple view geometry in computer
  vision}.\hskip 1em plus 0.5em minus 0.4em\relax Cambridge university press,
  2003.

\bibitem{KITTI}
A.~Geiger, P.~Lenz, and R.~Urtasun, ``Are we ready for autonomous driving? the
  kitti vision benchmark suite,'' in \emph{2012 IEEE Conference on Computer
  Vision and Pattern Recognition}.\hskip 1em plus 0.5em minus 0.4em\relax IEEE,
  2012, pp. 3354--3361.

\bibitem{ASN}
Y.~Zhang, S.~Khamis, C.~Rhemann, J.~Valentin, A.~Kowdle, V.~Tankovich,
  M.~Schoenberg, S.~Izadi, T.~Funkhouser, and S.~Fanello, ``Activestereonet:
  End-to-end self-supervised learning for active stereo systems,'' in
  \emph{Proceedings of the European Conference on Computer Vision (ECCV)},
  2018, pp. 784--801.

\bibitem{Cityscapes}
M.~Cordts, M.~Omran, S.~Ramos, T.~Rehfeld, M.~Enzweiler, R.~Benenson,
  U.~Franke, S.~Roth, and B.~Schiele, ``The cityscapes dataset for semantic
  urban scene understanding,'' in \emph{Proceedings of the IEEE conference on
  computer vision and pattern recognition}, 2016, pp. 3213--3223.

\bibitem{Depth_metric}
D.~Eigen, C.~Puhrsch, and R.~Fergus, ``Depth map prediction from a single image
  using a multi-scale deep network,'' in \emph{Advances in neural information
  processing systems}, 2014, pp. 2366--2374.

\bibitem{Tensorflow}
M.~Abadi, P.~Barham, J.~Chen, Z.~Chen, A.~Davis, J.~Dean, M.~Devin,
  S.~Ghemawat, G.~Irving, M.~Isard \emph{et~al.}, ``Tensorflow: A system for
  large-scale machine learning,'' in \emph{12th $\{$USENIX$\}$ symposium on
  operating systems design and implementation ($\{$OSDI$\}$ 16)}, 2016, pp.
  265--283.

\bibitem{RMSProp}
G.~Hinton, N.~Srivastava, and K.~Swersky, ``Neural networks for machine
  learning lecture 6a overview of mini-batch gradient descent,'' \emph{Cited
  on}, vol.~14, p.~8, 2012.

\end{thebibliography}
\bibliographystyle{IEEEtran}

%








\end{document}